\documentclass[10pt,twocolumn,letterpaper]{article}

\usepackage[pagenumbers]{cvpr} 

\usepackage{mathrsfs} 
\usepackage{overpic}   
\usepackage{siunitx}   
\usepackage{multirow}  
\usepackage{pifont}
\usepackage[flushmargin]{footmisc}
\newcommand{\cmark}{\ding{51}}  
\newcommand{\xmark}{\ding{55}} 

\newif\ifdraft
\drafttrue

\ifdraft
\DeclareRobustCommand{\raj}[1]{{\color{magenta}[\textbf{Raj:} #1]}}
\DeclareRobustCommand{\guan}[1]{{\color{purple}[\textbf{Guan:} #1]}}
\DeclareRobustCommand{\itai}[1]{{\color{orange}[\textbf{Itai:} #1]}}
\DeclareRobustCommand{\rana}[1]{{\color{cyan}[\textbf{Rana:} #1]}}

\DeclareRobustCommand{\TODO}[1]{\textbf{\color{red}[TODO: #1]}}

\else
\DeclareRobustCommand{\raj}[1]{}
\DeclareRobustCommand{\guan}[1]{}
\DeclareRobustCommand{\itai}[1]{}
\DeclareRobustCommand{\rana}[1]{}

\DeclareRobustCommand{\TODO}[1]{}

\fi

\definecolor{draftgrey}{rgb}{0.35,0.35,0.42}
\ifdraft
\newenvironment{draftnote}{\par\medskip\color{draftgrey}\small\noindent}{\par\medskip}
\else

\fi

\DeclareRobustCommand{\ourmethod}{DynaMesh}

\DeclareRobustCommand{\Mcal}{\mathcal{M}}   
\DeclareRobustCommand{\Tcal}{\mathcal{T}}   
\DeclareRobustCommand{\Ical}{\mathcal{I}}   
\DeclareRobustCommand{\Lcal}{\mathcal{L}}   
\DeclareRobustCommand{\Scal}{\mathcal{S}}   
\newcommand{\Fcal}{\mathcal{F}}

\definecolor{dynablue}{rgb}{0.0,0.21,0.96}
\definecolor{dynayellow}{rgb}{0.96,0.69,0.0}
\definecolor{dynagreen}{rgb}{0.55,0.70,0.50}

\definecolor{cvprblue}{rgb}{0.21,0.49,0.74}
\usepackage[pagebackref,breaklinks,colorlinks,allcolors=cvprblue]{hyperref}

\def\paperID{92} 
\def\confName{3DV\xspace}
\def\confYear{2027\xspace}

\title{\ourmethod{}: Dynamic 3D Texture Generation}

\author{
\vspace{1mm}
Raj Hansini$^{*}$ \hspace{0.9cm} Guan Chen$^{*}$ \hspace{0.9cm} Rana Hanocka \hspace{0.9cm} Itai Lang \\
University of Chicago \\
{\tt\small \{rajhansini, guanc, ranahanocka, itailang\}@uchicago.edu}
}

\begin{document}
\twocolumn[{
\renewcommand\twocolumn[1][]{#1}
\maketitle
\vspace{-1.9em}   
\begin{center}
    \centering
    \includegraphics[width=0.99\textwidth, trim=20 0 20 0, clip]{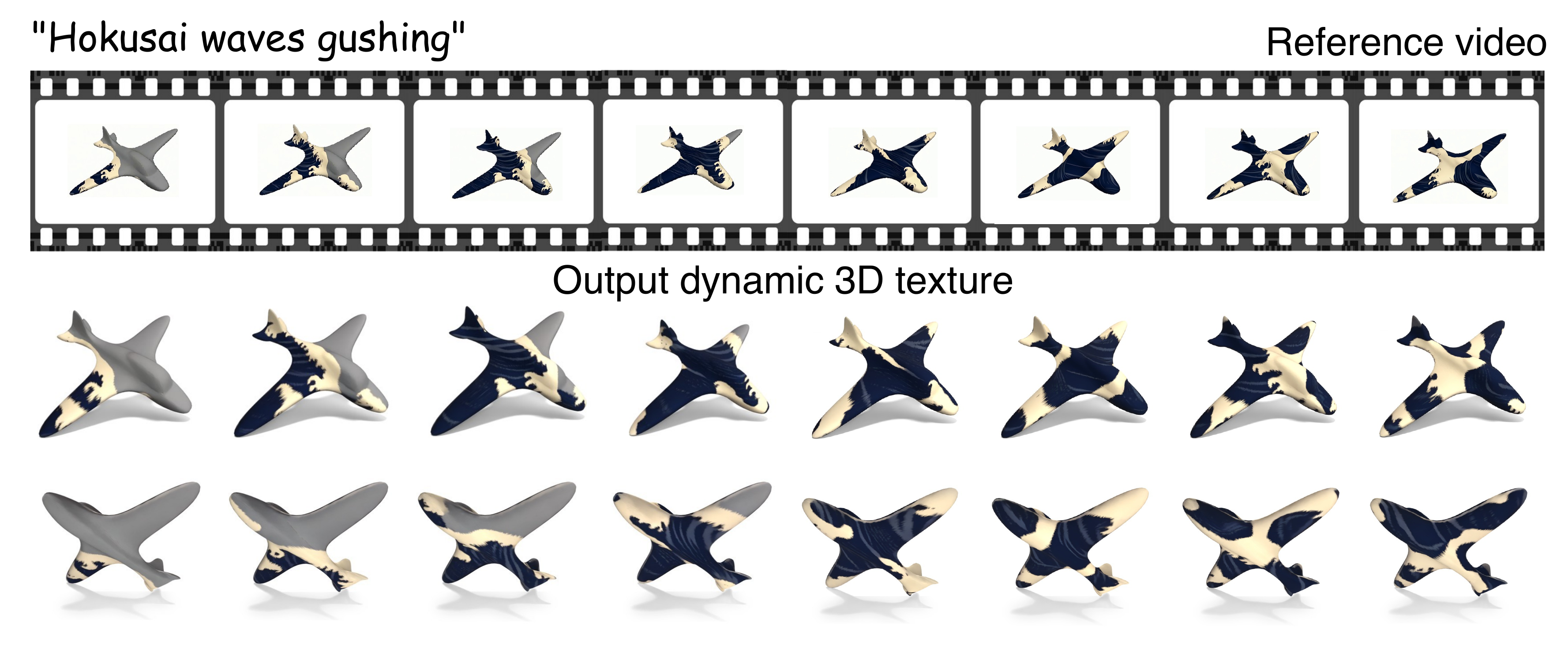}
    \vspace{-4mm}
    \captionof{figure}{\textbf{\ourmethod{} carries a video's evolving effect onto a 3D object as a texture that changes over time.} Given a reference video of the prompted effect unfolding on a render of the object (top), we produce a single mesh whose texture follows the video frame by frame (bottom, shown from a rotating viewpoint as time advances). The geometry stays fixed while the appearance evolves, so every moment of the sequence can be re-rendered from any viewpoint.}
    \label{fig:teaser}
\end{center}

}]

\def\thefootnote{*}\footnotetext{Equal contribution}

\begin{abstract}
We present \ourmethod{}, a dynamic texture generation method for 3D meshes. Given a textureless shape and a text prompt describing an effect, our method produces an appearance that evolves while the object's geometry remains unchanged. Previous works on dynamic 3D content generation have focused on motion, where an object's geometry and position change while keeping its appearance the same. Methods on texture generation sit on the other side of the problem, painting appearance onto a shape as a fixed surface property and not as an evolving process. Neither addresses a visual effect that propagates on a 3D object.

A natural route consists of two generators: a video model that shows the effect from a single view, and an image-to-3D generator that lifts each frame to 3D. However, the latter has no notion of time, so running it per video frame produces a sequence that flickers, loses effect details, and yields a different mesh at every video frame. Our method addresses these failures by conditioning a video model on a render of the mesh and the prompt to obtain a reference video, then running a frozen image-to-3D generator on the video with two changes. The conditioning of each frame is blended over a temporal window, and low-rank adapters are fit per shape to restore the lost details. The mesh is encoded once for the whole sequence, so geometry is constant by construction, and the output is a single mesh with a texture per frame. Applied to various objects and effects, \ourmethod{} substantially improves over recent video-to-4D and texturing methods, and can generalize its temporal effect to different shapes never seen during training. Our project page is at \url{https://threedle.github.io/dynamesh/}.
\end{abstract}
\section{Introduction}
\label{sec:intro}

\begin{figure*}[!t]
    \centering
    \includegraphics[width=0.99\textwidth, trim=10 0 10 0, clip]{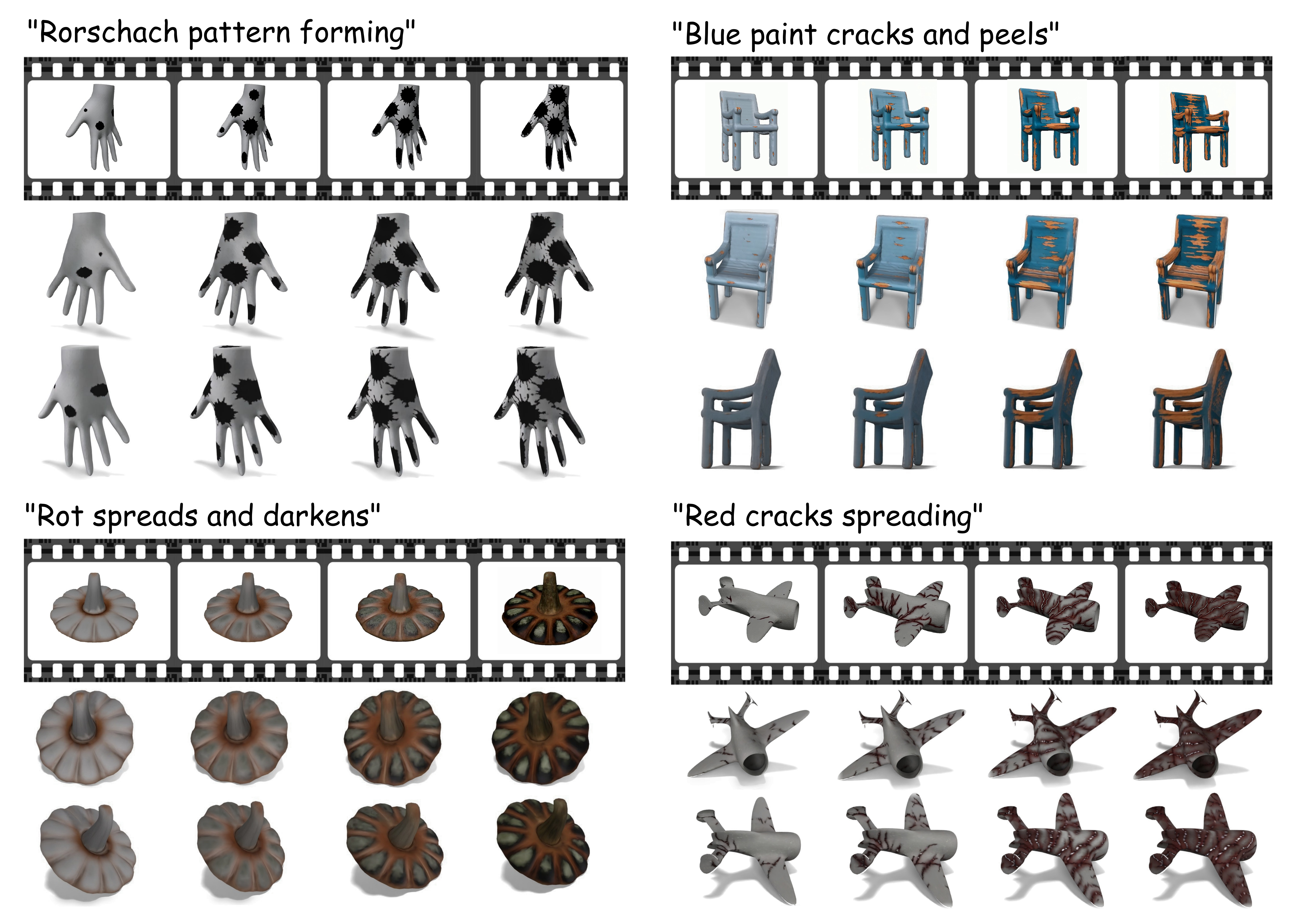}
    \vspace{-0.2cm}
    \caption{\textbf{Gallery of results.} Four objects driven by four different effects. Each panel shows the driving video as a film strip and, below it, our output rendered from two novel viewpoints. The texture remains coherent even at viewpoints unseen in training.}
    \vspace{-0.2cm}
    \label{fig:gallery}
\end{figure*}

Appearance change is an everyday phenomenon: ink spreading through paper, rust developing across a railing, frost blooming over a window, or paint peeling off a chair. Each of these examples is a process that propagates over the surface and changes its appearance over time. While the graphics literature has studied static object appearance~\cite{efros1999texture, turk2001texture, lefebvre2006appearance, michel2021text2mesh, richardson2023texture, decatur2025pixbrush, zhang2026mv2uv} and object motion~\cite{terzopoulos1987elastically, witkin1988spacetime, sumner2004deformation, vlasic2008articulated, pumarola2021dnerf, park2021nerfies, luiten2024dynamic} at length, giving a 3D asset an evolving surface texture remains largely unexplored.


Recent generative models each capture part of this problem. Video diffusion models~\cite{blattmann2023stable, yang2024cogvideox, wan2025wan} synthesize such phenomena convincingly, but their output is a flat sequence showing \textit{a single view}, with no underlying mesh. A growing body of work in dynamic 3D generation lifts video priors into four dimensions, producing objects and scenes that change over time from text, an image, or a single video~\cite{singer2023textto4d, bahmani20234dfy, zhao2023animate124, jiang2023consistent4d, ren2023dreamgaussian4d, ren2024l4gm, yao2025sv4d2, bahmani2024tc4d}. In all of them, however, the driving change is \textit{motion}. The target object transforms while its appearance is settled at the outset and carried along by the moving geometry.

In parallel, image-to-3D models~\cite{hong2023lrm, zhao2025hunyuan3d, xiang2024structured, xiang2025trellis2} turn one picture into a detailed textured mesh, an asset that can be viewed and edited from any direction but has a \textit{static appearance}. The same can be said for texture generation methods with a mesh as input, along with either text~\cite{chen2023text2tex, cao2023texfusion, xiang2024maketexture}, an image~\cite{zeng2023paint3d, zhang2026mv2uv}, or several modalities~\cite{yan2025flexpainter}. To date, the generation of dynamic textures for 3D shapes has not been sufficiently explored, with existing efforts confined to repeating patterns on a mesh~\cite{pajouheshgar2024meshnca} or to texturing an object whose motion is given~\cite{bao2024tex4d}. None of the model classes above offers the phenomena we look for, a surface seen from anywhere that changes over time. This raises the interesting question we pursue: \textit{can we generate a dynamic appearance for a 3D object?}

The difficulty is that this problem cannot be solved by simply pairing a video model with a 3D generator. A 3D generator has no temporal aspect as it simply maps one image to one object with no context of what came before. Running it frame by frame thus produces a sequence that flickers, since even near-identical frames will look different. Simply projecting the video onto the mesh also fails, as the video only shows one view of the object. Learning the task directly is equally out of reach without datasets of 3D objects with time-varying appearance. 

We thus introduce \ourmethod{}, a method for generating dynamic textures on 3D meshes. Given a mesh and a text prompt describing a temporal effect, we render a single view of the object and ask a frozen video model~\cite{kling2026} for that effect unfolding from the given view. Because the camera and object are held still, the returned video focuses only on appearance change. Then, we take an image-to-3D generator built for static objects~\cite{xiang2025trellis2} and temporally blend its input frames to tackle flicker, allowing the effect seen from one view to be learned as appearance evolving over the entire surface. The result is a single mesh with one triangulation and a texture per frame, an asset that can be rendered from any viewpoint at any moment of the effect.

\begin{figure}[!t]
    \centering
    \includegraphics[width=1.00\linewidth, trim=10 10 20 10, clip]{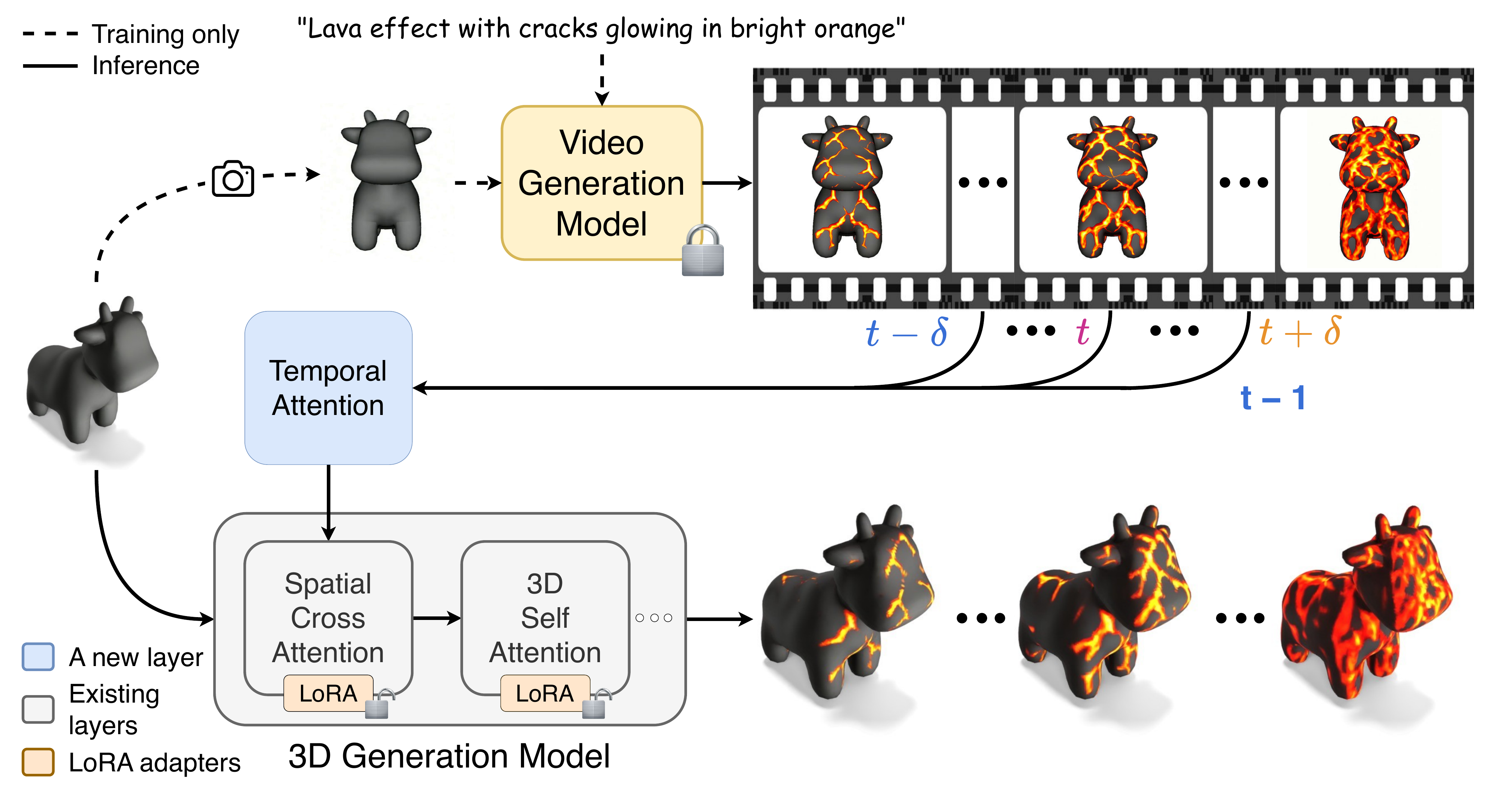}
    \vspace{-0.5cm}
    \caption{\textbf{System overview.} A render of the input mesh, together with a text prompt describing the effect, is passed to a video generation model, which returns the reference video for our pipeline. The 3D generator then produces the textured object at every frame. Our temporal cross-attention (the new module) conditions each frame's generation on a window of neighboring video frames. Low-rank adapters on the attention modules are fit per scene to recover detail. Geometry is shared across frames, so the output is one mesh whose texture evolves with the video.}
    \vspace{-0.2cm}
    \label{fig:overview}
\end{figure}

\ourmethod{} contributes in two ways. First, the generator inputs are temporally blended. Rather than conditioning at each time step on its own video frame, we blend the conditioning over neighboring frames by augmenting the 3D generator with temporal attention. Second, we fit low-rank adapters per scene against the driving video, recovering the fine detail that a generator constrained toward 3D consistency would otherwise lose. The mesh is encoded as input for the generator and shared by every frame, so the object's identity is preserved by construction and only the texture varies. 

\begin{figure*}[!t]
    \centering
    \includegraphics[width=0.99\textwidth]{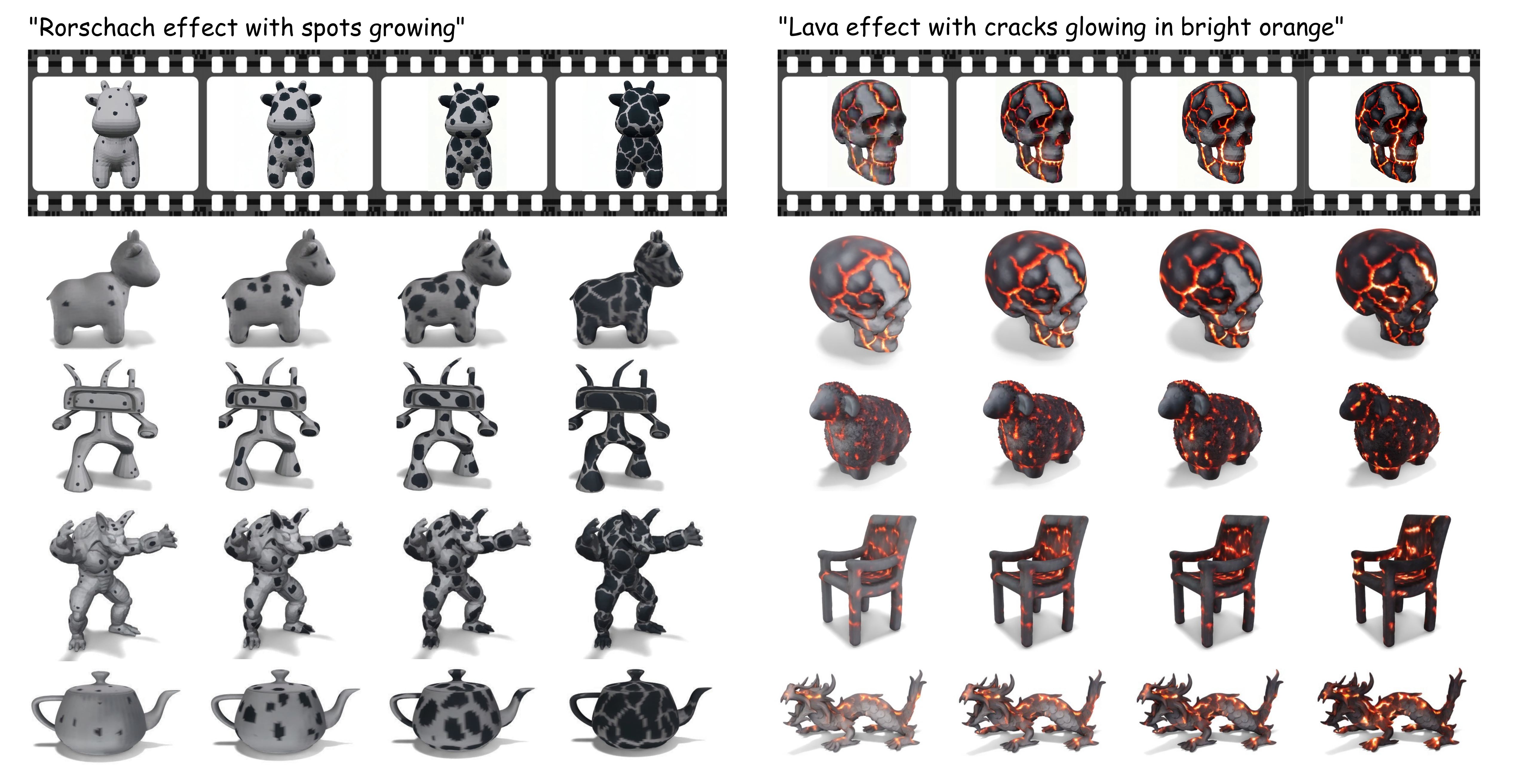}
    \vspace{-0.3cm}
    \caption{\textbf{Generalization.} Per-scene adapters fit on a reference video are applied, without any retraining, to meshes never seen during fitting. Both effects clearly propagate on the new shapes according to the video. The adapters change only how appearance is generated, and each object keeps its own geometry, so the same spreading blots or lava cracks work even on different shapes.}
    \vspace{-0.3cm}
    \label{fig:same-effect-diff-shape}
\end{figure*}

We show that \ourmethod{} is highly flexible. Across a variety of meshes and prompts, it produces diverse temporal effects that are coherent from viewpoints unseen in the reference video (see \cref{fig:gallery}). The adaptation also generalizes beyond the object it was trained on, as seen in \cref{fig:same-effect-diff-shape}, where the texture transfers across unseen shapes without requiring retraining. 
 
\section{Related work}
\label{sec:relatedwork}

\paragraph{Video Generation.}
Diffusion-based video models~\cite{ho2022video, blattmann2023stable, yang2024cogvideox, wan2025wan} synthesize short clips with convincing dynamics, and their image-to-video variants~\cite{blattmann2023stable, xing2023dynamicrafter} animate a single still image and largely preserve its content.

\begin{figure}[!t]
    \centering
    \includegraphics[width=\columnwidth]{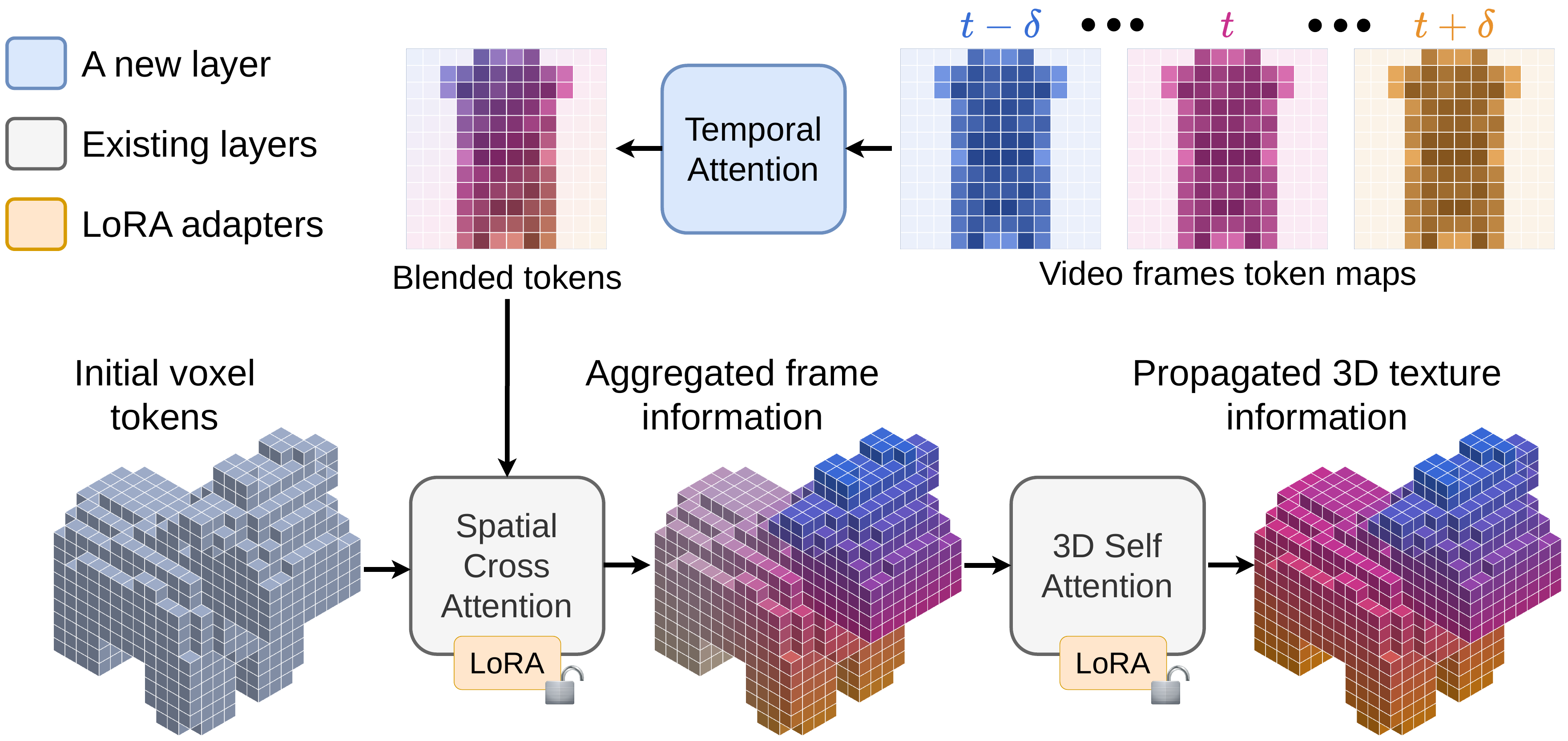}
    \vspace{-0.5cm}
    \caption{\textbf{Method overview.} The token maps of an eleven-frame window are blended by a temporal attention layer. The blend conditions the generator's spatial cross-attention, which writes the frame's appearance onto the voxel tokens of the input mesh, and the 3D self-attention then propagates that appearance across the surface. Every weight except the rank-4 adapters on those two attentions is frozen.}
    \vspace{-0.3cm}
    \label{fig:technique}
\end{figure}

Consistency in these models comes from denoising the clip jointly, where all frames share a single sampling trajectory. A per-frame generator has no such coupling. Even with the noise held fixed, the map from image to 3D output carries no stability guarantee, so two near-identical frames can land on visibly different reconstructions.

A substantial body of work aims to address the temporal consistency of such generations. Sharing the noise across frames and extending self-attention across frames suppress flicker in zero-shot video synthesis~\cite{khachatryan2023text2video}, and anchoring generation to 3D-consistent signals such as rendered correspondences or tracking information stabilizes the result further~\cite{cai2023generative, gu2025diffusion}.

Inspired by these works, we fix the noise seed across frames, so every frame is denoised from the same noise sample. We then blend each frame's conditioning tokens with those of its neighbors, so temporal context enters the generator rather than being imposed on its output. Since the generator is driven by a fixed mesh shared across frames, the result is a dynamic texture that can be rendered from any viewpoint.

\paragraph{Generative 3D Modeling.}
Early text-to-3D methods distill score gradients from pretrained 2D diffusion models into a per-scene 3D representation~\cite{poole2022dreamfusion, lin2023magic3d, chen2023fantasia3d} at the cost of slow per-scene optimization. Another line of work makes the 2D prior itself multi-view aware. Zero-1-to-3~\cite{liu2023zero} fine-tunes an image diffusion model conditioned on relative viewing angles to synthesize novel views of an input image, MVDream~\cite{shi2023mvdream} and Wonder3D~\cite{long2023wonder3d} generate mutually consistent view sets, and MV-Adapter~\cite{huang2024mvadapter} reaches the same goal by attaching adapters to a frozen 2D generator.

In contrast, feed-forward 3D generation models remove per-scene optimization altogether, generating a 3D representation directly from an image~\cite{hong2023lrm, zhao2025hunyuan3d}. Among these, TRELLIS~\cite{xiang2024structured, xiang2025trellis2} generates structured latents, a sparse grid of voxels carrying latent features, decoded into an explicit textured mesh. We use it as our backbone for its image conditioning, its feed-forward speed, and its separation of the voxel structure from the features it carries, which is what lets us hold geometry fixed while the texture evolves. TRELLIS handles 3D consistency \textit{per conditioning frame}, while our problem is consistency \textit{along the frames of the reference video}. Applied independently per frame, it produces temporal inconsistency, the challenge we address in this work. 

\paragraph{3D Texturing.}
Separate from generating a 3D shape from scratch, texturing an existing one is its own line of research. Text2Mesh~\cite{michel2021text2mesh} optimizes vertex colors and displacements against a vision-language objective. TEXTure~\cite{richardson2023texture} and Text2Tex~\cite{chen2023text2tex} paint a mesh view by view with a depth-conditioned diffusion model, inpainting each new view against the already-textured regions. Other variants transfer the texture of an image onto a given mesh~\cite{yeh2024texturedreamer}, while 3D Paintbrush~\cite{decatur20233d} and 3D PixBrush~\cite{decatur2025pixbrush} localize texture edits to a region of the surface.

Recent work moves beyond generative image models and borrows a temporal video prior, treating an orbit around the object as a video so that its smoothness carries over to the texture~\cite{kang2025consistent}. TRELLIS~\cite{xiang2024structured, xiang2025trellis2} folds texturing into the generation process itself, decoding per-voxel appearance features into PBR surface attributes. As mentioned above, none of these are temporally consistent when run per video frame. Our method treats texture as a \textit{sequence}, generating every frame from the same frozen 3D generator~\cite{xiang2025trellis2} under our novel temporal conditioning mechanism.

\paragraph{Dynamic 3D Generation.}
Work extending generation to 4D has focused on motion. Text- and video-conditioned methods deform 3D representations over time~\cite{singer2023textto4d, jiang2023consistent4d, bahmani20234dfy, ren2023dreamgaussian4d, bahmani2024tc4d} or generate multi-view videos to reconstruct the moving object~\cite{yao2025sv4d2, ren2024l4gm}. Along this line, the \textit{texture is essentially fixed}: it is painted once at the start and carried along with the deformation, and the modeling focuses on the consistency of the moving geometry.

Tex4D~\cite{bao2024tex4d} tackles the inverse problem. Instead of animating the shape under a fixed texture, it starts from an already-animated mesh sequence and uses a video diffusion model to produce textures that remain consistent while the geometry deforms. A method closer to our setting is MeshNCA~\cite{pajouheshgar2024meshnca}, which synthesizes animated textures on a static mesh with neural cellular automata. Its texture moves as a repeated pattern and does not progress, and the motion comes from a hand-chosen field.  

The classical literature achieved appearance change over time by other means. Stochastic processes model it without any surface~\cite{doretto2003dynamic}, and weathering work simulates or captures \textit{specific} phenomena such as patinas or decay on static objects~\cite{dorsey1996modeling, gu2006time}. To our knowledge, no prior method generates an open-ended, directed appearance evolution on a given 3D shape from a \textit{general} generative video prior.

\section{Method}
\label{sec:method}

Given a textureless mesh $\Mcal$ and a text prompt $Y$ describing an evolving surface effect, our goal is to produce a sequence of textures $\Tcal_1,\dots,\Tcal_T$ on the fixed triangulation of $\Mcal$, such that rendering $\Mcal$ with $\Tcal_t$ from any viewpoint shows the effect at time $t$. We approach this in two stages. First, we render $\Mcal$ from a single view and pass that render with $Y$ to a video model~\cite{kling2026}, which returns $\Ical_1,\dots,\Ical_T$ depicting the effect from that view (\cref{fig:overview}). Second, we drive an image-to-3D generator $\mathcal{G}$~\cite{xiang2025trellis2} with these frames to lift the observed effect onto the whole surface.

The na\"{\i}ve route for generating the temporal appearance in 3D is to apply the 3D generator to each frame. This fails in three ways. First, consecutive 3D texture outputs flicker since each is generated in isolation. Second, the effect washes out, as the generator's prior pulls fine detail toward the textures it was trained on. Lastly, each frame yields a \emph{different} mesh, so there is no single object to texture. Our method, \ourmethod{}, addresses each failure in turn by coupling neighboring frames for temporal attention, adapting the generator per scene \wrt the reference video to restore the lost detail, and encoding the given mesh once and sharing it across frames to fix geometry by construction, as illustrated in \cref{fig:technique}.

\ourmethod{} builds on a 3D generator that separates appearance and geometry~\cite{xiang2025trellis2}. A shape encoder $\mathcal{E}$ maps an input mesh to a structured latent $\Scal$, a set of occupied voxels each carrying a geometry feature. A flow model then predicts per-voxel appearance features conditioned on $\Scal$ and on an input image encoded into tokens by a pretrained feature extractor~\cite{simeoni2025dinov3}. At each block, spatial cross-attention writes the image tokens onto the voxels, and 3D self-attention propagates appearance between the voxels. A decoder $\mathscr{D}$ then expands those features into PBR surface attributes on the mesh that $\Scal$ encodes.

We run this generator once per video frame and address the aforementioned failures by intervening at three points: (1) each conditioning token is blended with the tokens at the same spatial position in neighboring frames before the generator attends to them, so consecutive frames share context, and the operation is parameter-free (\cref{subsec:mcfm}); (2) low-rank adapters on the spatial cross-attention and the 3D self-attention are fit to the reference video, so the generator reproduces this effect, not the generic texture its prior prefers (\cref{subsec:lora}); (3) the input mesh is encoded by the 3D generator once, so every frame textures the same geometry.

The adapters are trained with an image-space loss between a render of the generated texture on $\Mcal$ and the corresponding video frame. The structured latent $\Scal$ and the sampling noise are shared across all frames, so every frame is generated on the same geometry from the same noise.

\subsection{Temporal Frame Attention}
\label{subsec:mcfm}

\begin{figure}[!t]
    \centering
    \includegraphics[width=\columnwidth]{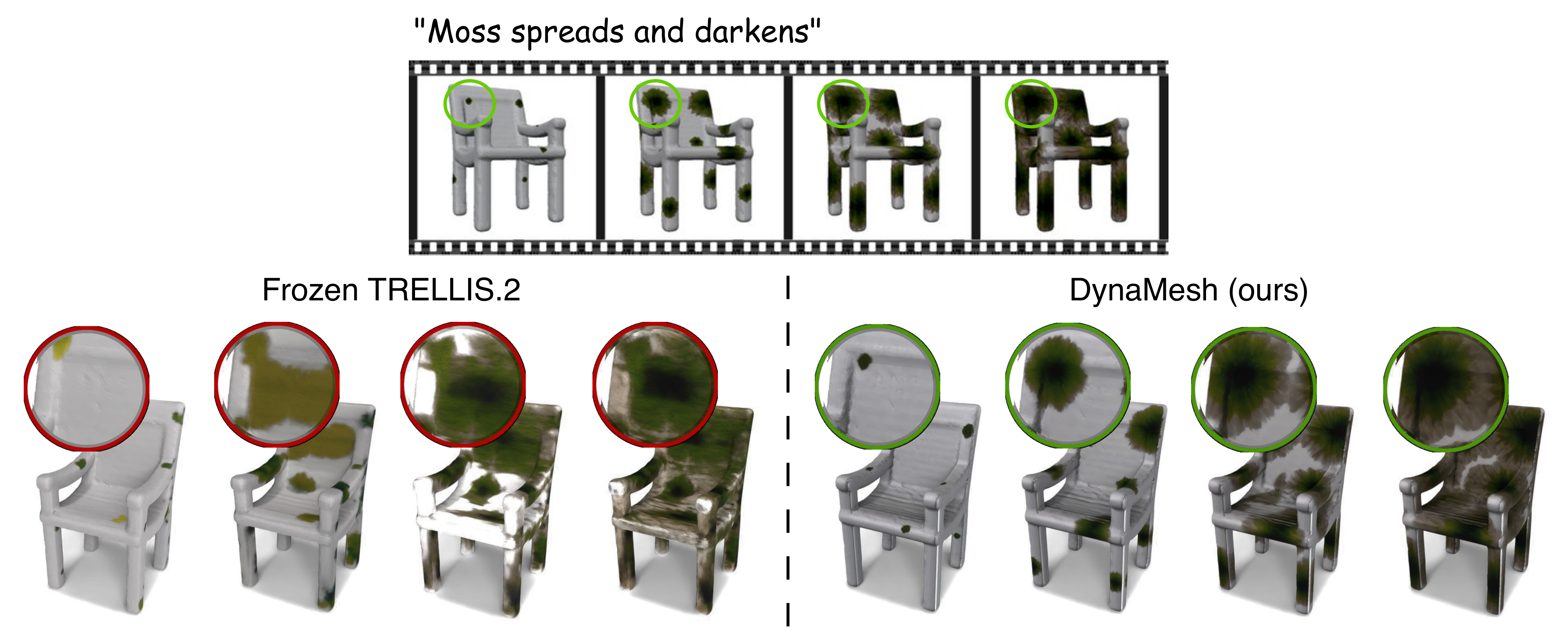}
    \vspace{-0.5cm}
    \caption{\textbf{Temporal coherence.} A chair with moss spreading over its surface is shown at four time frames. The insets magnify the same patch of the backrest. Frozen TRELLIS.2 changes the texture abruptly from one moment to the next (flickers), while our method keeps the moss growing smoothly, as in the video.}
    \vspace{-0.3cm}
    \label{fig:flicker}
\end{figure}

A frozen image-to-3D generator run independently over a video decides each frame in isolation, and the resulting temporal texture flickers (\cref{fig:flicker}). The cause is not the sampling noise. We hold the sampling noise fixed, and $\Scal$ is a deterministic encoding of an unchanging mesh, so the conditioning tokens are the only input that varies with $t$. Consecutive frames of the reference video are nearly identical, yet their reconstructions differ visibly. Running a generator built for static objects on a smoothly varying sequence exposes a property the single-image setting hides: small changes in its conditioning produce large changes in its output.

Removing this flicker requires no learned parameters. All frames are encoded by the same frozen DINOv3 encoder, so their tokens lie in a common space. We therefore blend each frame's conditioning tokens with those of its neighbors before the generator attends to them, weighting each neighbor by its similarity in that space; the weights in \cref{eq:blend} are softmax-normalized similarities.

Let $z_t$ denote the conditioning tokens of frame $\Ical_t$, computed once by the frozen feature extractor. Before the 3D generator's cross-attention sees them, we replace each token by a blend over a temporal window $\mathcal{W}$ centered at $t$,
\begin{equation} \label{eq:blend}
    \tilde{z}^{\,i}_t \;=\; \sum_{\delta \in \mathcal{W}} \alpha^{\,i}_{\delta}\, z^{\,i}_{t+\delta},
\end{equation}
\begin{equation} \label{eq:softmax}
   \alpha^{\,i}_{\delta} \;=\; \mathrm{softmax}_{\delta \in \mathcal{W}}\!\left( \frac{\langle z^{\,i}_t,\, z^{\,i}_{t+\delta} \rangle}{\sqrt{D}} \right),
\end{equation}
where $i$ indexes token positions and $D$ is the token dimension. Each position attends over the same position in the neighboring frames, with the current frame as the query. Where consecutive frames agree, neighboring tokens score highly and are averaged, suppressing frame-to-frame noise. Where the effect meaningfully changes, the current frame dominates its own weight and the change passes through.

Restricting each token to its own position across time is important since this makes the conditioning spatially anchored, meaning each token represents the same region of the object across every frame. The mechanism is parameter-free, since the weights are computed directly from the tokens (\cref{eq:softmax}).

\subsection{3D Generator Adaptation}
\label{subsec:lora}

\begin{figure}[!t]
    \centering
    \includegraphics[width=\columnwidth]{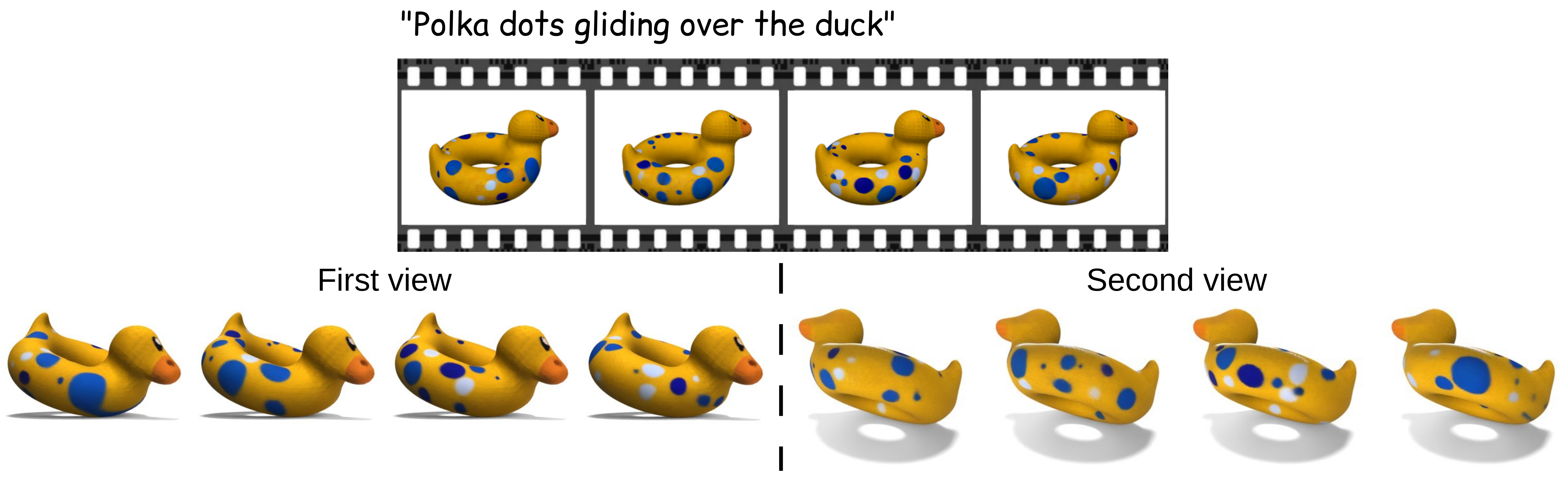}
    \vspace{-0.5cm}
    \caption{\textbf{Making an existing static texture dynamic.} Our method can operate on objects with a given texture and make it change over time, as shown for the moving spots over the duck.}
    \vspace{-0.3cm}
    \label{fig:existing_texture}
\end{figure}

Temporal conditioning stabilizes the sequence, but stability is insufficient. The frozen generator still loses detail relative to the reference video, so the effects come out low-fidelity. This is because the effect lies outside what the generator's prior reproduces from a conditioning image, so we address this with trainable parameters. We keep them minimal, fitting low-rank adapters~\cite{hu2021lora} per shape against the driving video while the rest of the generator's parameters stay frozen. We can write
\begin{equation} \label{eq:lora}
\begin{aligned} 
      W' &=\; W + \Delta W, \\ \Delta W &=\; \frac{\alpha}{r}\, BA
\end{aligned}
\end{equation}
where $W \in \mathbb{R}^{d \times k}$ is a frozen projection, $B \in \mathbb{R}^{d \times r}$ and $A \in \mathbb{R}^{r \times k}$ are trained. The update is scaled by $\alpha / r$ following~\cite{hu2021lora}, with $\alpha = 4$. We use $r = 4$, and since $r \ll \min(d, k)$, the update is low rank. Across all adapted layers, the updates amount to less than a tenth of a percent of the generator's parameters.

The generator offers two locations, with different roles, for the learned low-rank adapter. The cross-attention projections carry image tokens onto the voxels, so adapting them changes what the generator reads from each frame. The self-attention shares information between voxels, so adapting it changes how appearance propagates over the object. We adapt the attention projections in every block, and the full formulas are in the supplementary material.

The remaining question is what to fit the adapters to. The video frame cannot serve as the training target, because while the object's \textit{position} and \textit{pose} are fixed via our prompt, the effect still changes its \textit{outline}. A pixel loss against the raw frame would penalize the adapter for a shape difference as though it were a texture error. 
We instead build a target per frame that matches the mesh silhouette by construction, taking color from the video where the two agree and filling the remaining rim, 0.6\% to 2.2\% of pixels, from the nearest valid neighbor. The loss is an $L_1$ penalty between the render and the target with a small perceptual term, computed inside the mesh silhouette.

Writing $R_t$ for our render at frame $t$, $\widehat{\Ical}_t$ for the constructed target, and $M_t$ for our mesh silhouette, the per-frame objective is
\begin{equation} \label{eq:loss}
    \Lcal_t = \frac{1}{3|M_t|} \sum_{p \in M_t} \big\| R_t(p) - \widehat{\Ical}_t(p) \big\|_1 + \lambda \, \mathrm{LPIPS}_{\square}\big(R_t, \widehat{\Ical}_t\big),
\end{equation}
where $\|\cdot\|_1$ sums over color channels, $\mathrm{LPIPS}_{\square}$ is evaluated on both images cropped to the bounding box of $M_t$, and $\lambda = 0.1$.

We train the adapters on the mean of $\Lcal_t$ over the video. Each training step runs the generator to a random point, then applies a single flow step with gradients enabled, decodes and renders the appearance, and updates the adapters only. Because each step stops at a different noise level, the adapters are supervised across the full range over training.

\subsection{Preserving the Geometry Across Frames}
\label{subsec:geometry}

Methods that generate shape per frame return a different mesh each time, with no correspondence between frames, so no UV layout, rig, or edit survives the sequence.

We use the 3D generator to encode the input mesh once. The shape encoder of the generator maps it deterministically to the structured latent $\Scal$, the only place geometry enters the generator, and the same $\Scal$ conditions appearance generation at every frame. We define $\Fcal_{W'}(\epsilon, \Scal, \tilde z_t)$ as the full denoising run of the adapted flow under the adapted weights $W'$ from \cref{eq:lora}, started from the shared noise $\epsilon$. Writing $\mathcal{E}$ for the shape encoder and $\mathscr{D}$ for the decoder, the full pipeline is
\begin{equation} \label{eq:pipeline}
\begin{aligned}
    \Scal \;&=\; \mathcal{E}(\Mcal), \\
    \Tcal_t \;&=\; \mathscr{D}\!\big(\Scal,\; \Fcal_{W'}(\epsilon,\, \Scal,\, \tilde{z}_t)\big).
\end{aligned}
\end{equation}     
The only input that varies with $t$ is the blended conditioning $\tilde{z}_t$, so appearance is the only thing that can change over time, and geometry is constant by construction.
\section{Experiments}
\label{sec:experiments}


We evaluate our dynamic 3D texture generation method across a variety of axes and against existing methods. \Cref{subsec:properties} shows the range of effects \ourmethod{} produces and its temporal stability. \Cref{subsec:transfer} demonstrates that a fitted effect carries over to shapes never seen while training the adapters. \Cref{subsec:evaluation} compares against baselines, ablates the method's components, and discusses limitations.

\begin{table}
\centering
\small
\setlength{\tabcolsep}{4.5pt}
\begin{tabular}{l cc cc}
\toprule
& \multicolumn{2}{c}{Unseen views} & \multicolumn{2}{c}{Fidelity \vs \ GT}
\\
\cmidrule(lr){2-3} \cmidrule(lr){4-5}
Method & Flicker\,$\downarrow$ & Accel.\,$\downarrow$
& PSNR\,$\uparrow$ & SSIM\,$\uparrow$ \\
\midrule
\textcolor{gray}{SV4D 2.0$^\dagger$~\cite{yao2025sv4d2}} & \textcolor{gray}{0.023} & \textcolor{gray}{0.035} & \textcolor{gray}{34.8} & \textcolor{gray}{0.963} \\
\textcolor{gray}{DG4D$^\dagger$~\cite{ren2023dreamgaussian4d}} & \textcolor{gray}{0.074} & \textcolor{gray}{0.142} &
\textcolor{gray}{13.3} & \textcolor{gray}{0.393} \\
L4GM~\cite{ren2024l4gm} & 0.007 & 0.011 & 21.6 & 0.728 \\
MeshNCA~\cite{pajouheshgar2024meshnca} & 0.018 & 0.033 & 7.7 & 0.105 \\
TRELLIS.2~\cite{xiang2025trellis2} & 0.018 & 0.027 & 12.1 & 0.303 \\
\ourmethod{} (ours) & \textbf{0.005} & \textbf{0.005} & \textbf{24.9} & \textbf{0.798} \\

\bottomrule
\end{tabular}
\caption{Quantitative comparison over 42 objects and effects and all 150 frames. Flicker and acceleration are per-step temporal errors, and PSNR/SSIM, fidelity metrics to the ground truth video, are computed in the supervised view against the video. We achieve better results than state-of-the-art methods. $^\dagger$SV4D~2.0 and DG4D are grayed out as their outputs have fewer frames and cannot be measured on the same frame grid, discussed more below.}
\label{tab:comparison}
\end{table}

\subsection{Implementation Details}
\label{subsec:impl}

For all results, we render the mesh from a single fixed camera and pass that render with the prompt to a video generation model~\cite{kling2026}, which produces the reference video in which the camera and the object are static and only the surface appearance evolves, for a prompt such as ``blue paint cracks and peels''. Our generator is TRELLIS.2~\cite{xiang2025trellis2} with conditioning tokens from a pretrained DINOv3 encoder~\cite{simeoni2025dinov3}. Multi-frame conditioning uses an eleven-frame window $[t{-}5, \dots,t, \dots, t{+}5]$. The per-shape LoRA adapters are rank-4 residuals on the query, key-value, and output projections of the spatial cross-attention and those of the 3D self-attention. They are fit for 30 epochs against the reference video using the objective of \cref{subsec:lora}, an $L_1$ term over the mesh silhouette plus a perceptual term weighted at $0.1$. The generator runs at $512^2$ and renders at $960^2$, on a single NVIDIA A40 GPU. Each sequence is 150 frames. Further training details are in the supplementary material.

\subsection{Properties of \ourmethod{}}
\label{subsec:properties}

\noindent\textbf{Effect diversity.} \Cref{fig:gallery} shows different objects, each guided by a different effect. Our dynamic textures vary in how the effect is organized on the surface. Rot on the pumpkin begins near the stem and darkens outward along the ribs, so the change has a source and a direction. The airplane behaves similarly under a different effect, with the red cracks propagating across the shape. The crack pattern on the chair is another prominent example of temporal consistency, as cracks that appear early stay anchored to the same surface points as the effect advances. The Rorschach pattern on the hand has no geometric anchor at all and is carried entirely by the video. Each example is shown from two \textit{unseen} views, and the effects are present and consistent on both.

\smallskip
\noindent\textbf{Textured inputs.} Our method also applies to meshes that are
already textured. \Cref{fig:existing_texture} shows a molded rubber duck float
whose own texture carries the dots; the prompted effect sets them in motion.
They glide around the ring while the head stays clear and the painted eyes,
beak, and molded shading hold their place, so only the pattern moves.

\subsection{Generalization}
\label{subsec:transfer}

\ourmethod{} generalizes across shapes because the adapter is kept separate from the shape encoding, so it captures the effect rather than the geometry. For example, \Cref{fig:same-effect-diff-shape} shows a Rorschach effect fitted on an animal transferring to a teapot, and lava cracks trained on a skull transferring to a chair, objects from different categories and with different geometry. These results suggest that our method learns the essence and dynamics of the effect.

\begin{figure}[!t]
    \centering
    \includegraphics[width=\columnwidth]{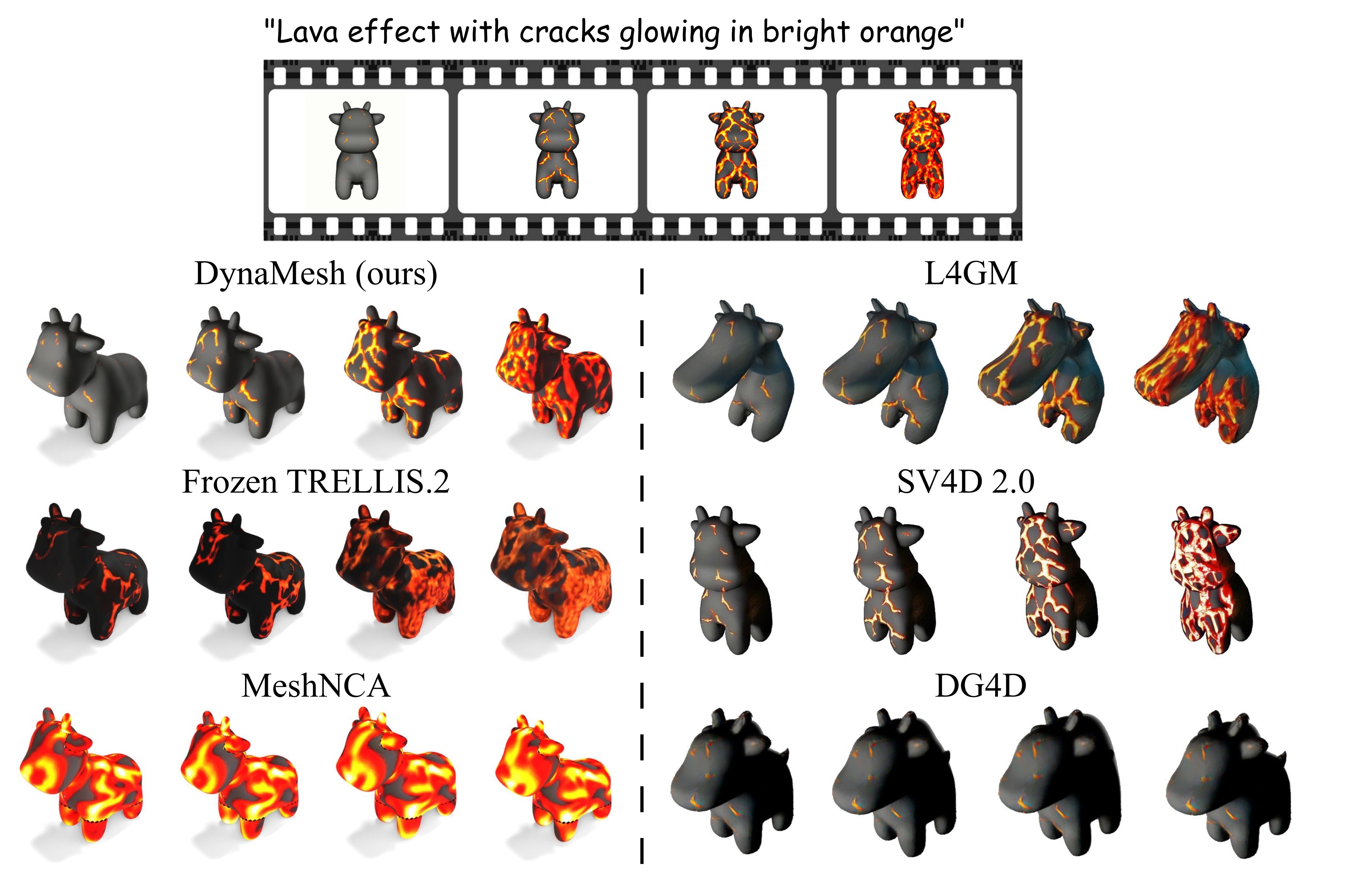}
    \vspace{-0.6cm}
    \caption{\textbf{Qualitative comparison.} We compare our method against the baselines on one reference video (top strip), with each method shown at four timesteps from a single unseen view. The baselines either flicker or degrade the geometry at that view. Only our method tracks the effect consistently and preserves the shape's geometry.}
    \vspace{-0.2cm}
    \label{fig:comparison}
\end{figure}

\subsection{Evaluation}
\label{subsec:evaluation}

\noindent\textbf{Protocol.} We compare against recent techniques for dynamic content generation~\cite{yao2025sv4d2, ren2023dreamgaussian4d, ren2024l4gm, pajouheshgar2024meshnca}, together with the frozen generator run per frame. All methods are rendered to $512 \times 512$ frame sequences at the supervised view and three unseen views.

For evaluating the temporal consistency of the results, we use two metrics, flicker and acceleration. Flicker measures the average per-frame change of the rendered sequence. It is defined as
\begin{equation} \label{eq:flicker}
    \mathrm{flicker} \;=\; \frac{1}{T-1} \sum_{t=2}^{T} \frac{1}{3|\Omega|} \sum_{p \in \Omega} \big\| R_t(p) - R_{t-1}(p) \big\|_1 ,
\end{equation}
where $R_t$ is the rendered frame at time $t$, $T$ is the number of frames, and $\Omega$ is the method's foreground, taken as the union of its masks at the first and last frames. Each method is measured over its own $\Omega$, since some baselines produce their own geometry. Acceleration measures abrupt texture changes across frames, defined as the second difference in pixel color:
\vspace{-0.1cm}
\begin{equation} \label{eq:accel}
\begin{split}
    \mathrm{accel} = {}& \frac{1}{T-2} \sum_{t=2}^{T-1} \frac{1}{3|\Omega|} \sum_{p \in \Omega} \\
    & \big\| R_{t+1}(p) - 2 R_t(p) + R_{t-1}(p) \big\|_1 .
\end{split}
\end{equation}

We also compute reconstruction metrics, PSNR and SSIM, at the supervised view against the frames. PSNR measures pixel-wise fidelity to the frame, and SSIM measures the preservation of local structure. Their formal definitions are provided in the supplementary.

\smallskip
\noindent\textbf{Baselines.} We compare against four methods, and the frozen generator run per frame as the baseline reference. SV4D~2.0~\cite{yao2025sv4d2} is a multi-view video diffusion model, L4GM~\cite{ren2024l4gm} is a feed-forward model for video-to-4D Gaussians, DreamGaussian4D (DG4D)~\cite{ren2023dreamgaussian4d} performs per-scene optimization for deformable Gaussians, and MeshNCA~\cite{pajouheshgar2024meshnca} synthesizes dynamic texture directly on the given mesh and is the closest prior method in output form. However, since it accepts only a textured image, we chose the last frame from the reference video as input to show the full effect.

\smallskip
\noindent\textbf{Qualitative comparison.} \Cref{fig:comparison} shows a reference video and one row per method. The frozen generator and MeshNCA receive the mesh and keep the object intact, but neither achieves the evolving appearance. The frozen generator reconstructs each frame independently, and MeshNCA sees only an image of the completed effect. The video-to-4D baselines follow the texture to some extent, but with low fidelity to the reference, since they must infer shape from a single view.

\smallskip
\noindent\textbf{Quantitative comparison.}  \Cref{tab:comparison} reports four evaluation quantities over 42 shapes and effects. Our method leads on every quantity among the comparable baselines: it more than doubles the frozen generator's reconstruction quality ($12.1 \rightarrow 24.9$\,dB, SSIM $0.30 \rightarrow 0.80$) and reduces its flicker at unseen views by nearly three-quarters and its acceleration by four-fifths.

L4GM is the closest baseline but remains $3.3$\,dB behind and flickers more at unseen views, while MeshNCA, which only has one textured image as input, fails on both reconstruction and temporal consistency. SV4D 2.0 and DG4D output 21 and 32 frames, respectively, so their values are computed on their own coarser frame grids, where each step spans ${\sim}7$ frames from the reference video. Their PSNR/SSIM columns are therefore not comparable to the rest, being computed on that coarser grid, and SV4D 2.0's are inflated further because its supervised view is a re-encoding of its own input.

\smallskip
\noindent\textbf{Ablation study.} In the supplementary, we ablate the method's components in texel space and show that each contributes to the final result. Furthermore, we visualize the reduction in flicker achieved by \ourmethod{}.

\begin{figure}[!t]
    \centering
    \includegraphics[width=\columnwidth]{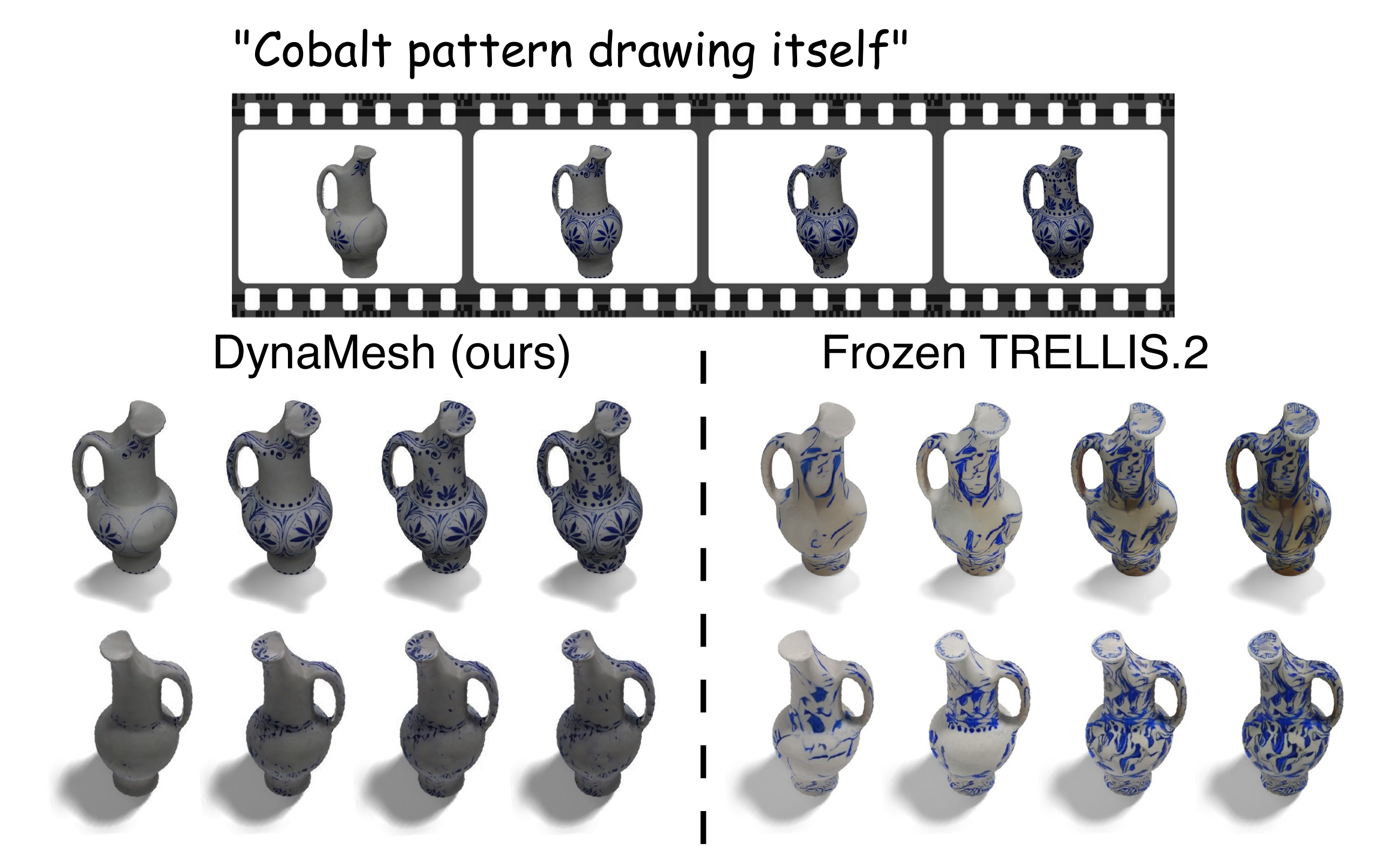}
    \vspace{-0.7cm}
    \caption{\textbf{Failure case.} \ourmethod{} relies on effects where the property of the pattern is local. Here, it replicates the local patterns (flowers, dots, etc.), but it does not recover them from unseen views, since the flowers and lines follow a global arrangement that a single view does not capture.}
    \vspace{-0.3cm}
    \label{fig:limitations}
\end{figure}


\smallskip
\noindent\textbf{Failure case.} Most of our effects are defined by local appearance, and those extend to unseen views. However, patterns defined by how their parts are arranged across the whole surface do not. For example, \Cref{fig:limitations} shows a vase that develops cobalt flower patterns. The supervised view came out as intended, but the pattern visibly loses detail on the unseen views. This is because the effect is defined not only by the local flowers, but also by the global arrangement of the lines with respect to each flower.
\section{Conclusion}
\label{sec:conclusions}

We introduced \ourmethod{}, a method that generates dynamic texture for a 3D object. Rather than retraining an image-to-3D generator to handle time, we leave it frozen and make two contributions. We blend the image conditioning tokens it attends to over a temporal window, and we fit a per-scene low-rank adaptation on its spatial cross-attention and 3D self-attention. Across a range of shapes and prompts, our method produces effects that progress in time, stay consistent from unseen viewpoints, and follow the reference video more closely than recent 4D generation techniques. Notably, our adapted generator trained on one shape can generalize the effect to other unseen objects.

Two limitations point to future work. The adaptation is fit once per video, so a new effect means a new fit. Training a single adaptation across multiple effects would remove that cost. The second is that geometry is fixed by construction, so effects that deform the surface are out of scope. A natural next step, which we intend to pursue, is to let the reference video drive deformation as well as appearance, giving a method that is temporally consistent in both texture and geometry.

{
    \small
    \bibliographystyle{ieeenat_fullname}
    \bibliography{references}
}

\clearpage
\appendix
\maketitlesupplementary
\providecommand{\cmark}{\ensuremath{\checkmark}}
\providecommand{\xmark}{\ensuremath{\times}}

The following sections provide more information regarding our dynamic 3D texture generation method. \Cref{sec:supp_webpage} points to our supplementary webpage, where every result is shown as a video. \Cref{sec:supp_additional} adds further objects and effects, a comparison on an object with thin separated parts, flicker measured at each time step, and an effect continued into a second clip. \Cref{sec:supp_judge} scores the unseen views with a vision-language model, which the image metrics cannot reach. \Cref{sec:supp_ablation} ablates the components, the window width, and the choice of temporal operator. Finally, \Cref{sec:supp_impl,sec:supp_metrics,sec:supp_baselines} give the settings, the metric definitions, and the baseline configurations.

\section{Supplementary Webpage} \label{sec:supp_webpage}

Our supplementary webpage provides the results of the paper as videos, each next to its reference video and shown in the supervised and unseen views. It includes the generalization to unseen shapes, the comparison against every baseline, and the failure case.

\section{Additional Results} \label{sec:supp_additional}

\begin{figure*}[!t]
    \centering
    \includegraphics[width=\textwidth]{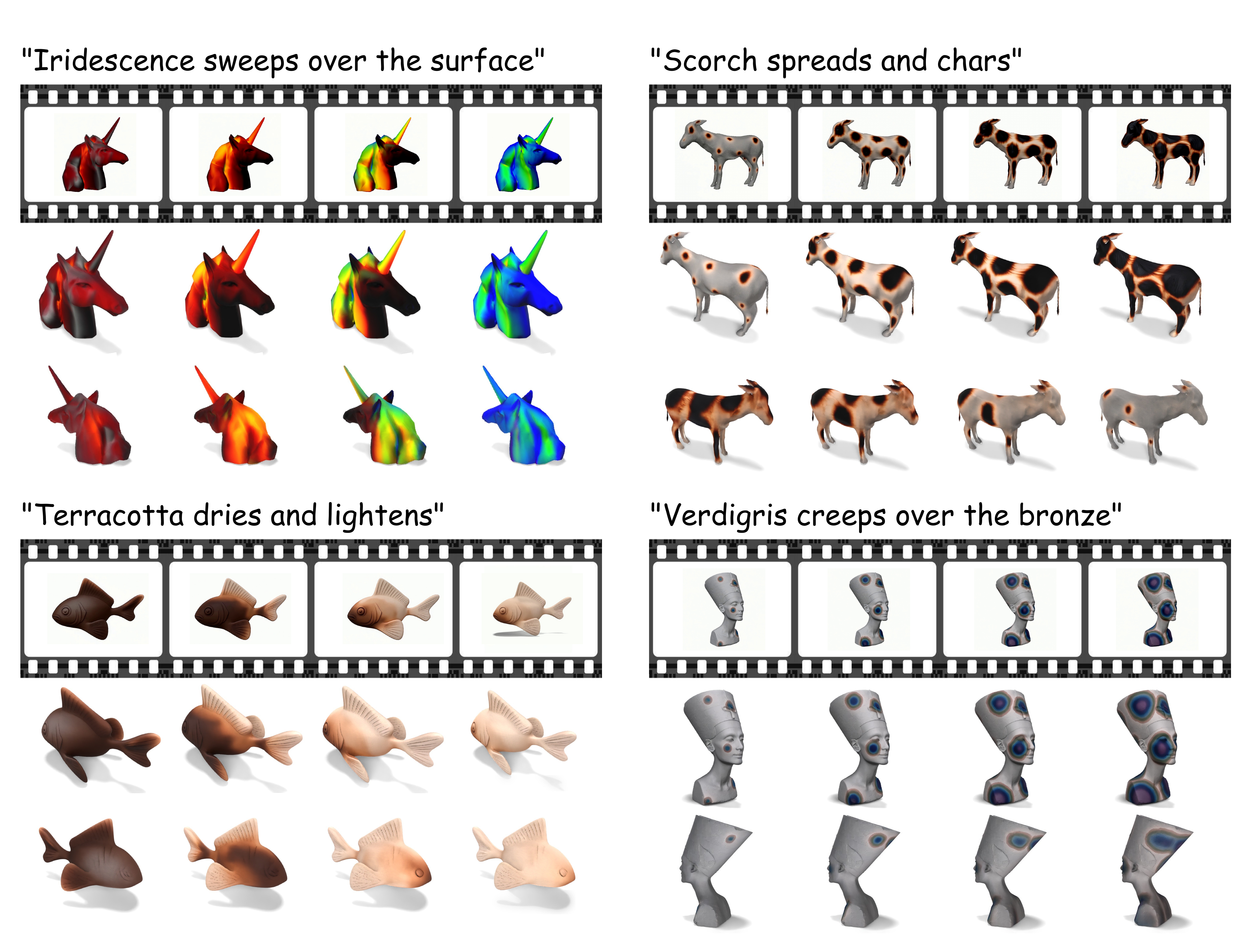}
    \vspace{-0.5cm}
    \caption{\textbf{Additional gallery.} DynaMesh generates a different temporal effect for each object, and the effect holds at an unseen viewpoint despite being fit to a single camera.}
    \vspace{-0.3cm}
    \label{fig:gallery_supp}
\end{figure*}

\noindent \textbf{Gallery.} The four effects in \cref{fig:gallery_supp} imitate different physical processes, from a surface oxidizing or charring to one drying out or turning iridescent, and each is produced under the settings of \cref{tab:supp_settings}.

\begin{figure*}[!t]
    \centering
    \includegraphics[width=\textwidth]{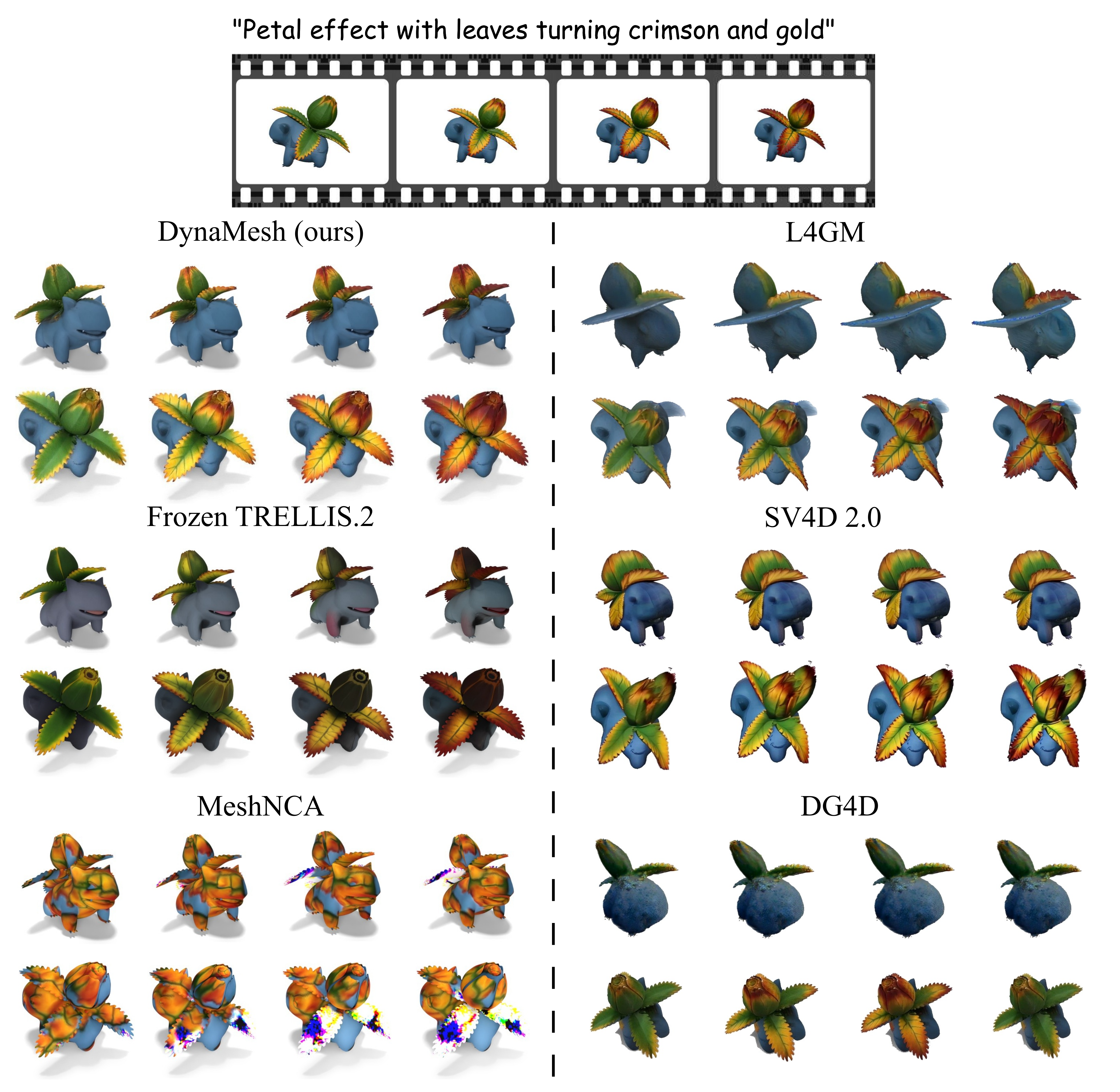}
    \vspace{-0.5cm}
        \caption{\textbf{Qualitative comparison.} Given the same reference video, the baselines either lose the effect or distort the geometry at the unseen viewpoint, whereas DynaMesh preserves both.}

    \vspace{-0.3cm}
    \label{fig:comparison_supp_a}
\end{figure*}

\begin{figure}[t!]
\centering
\includegraphics[width=\columnwidth]{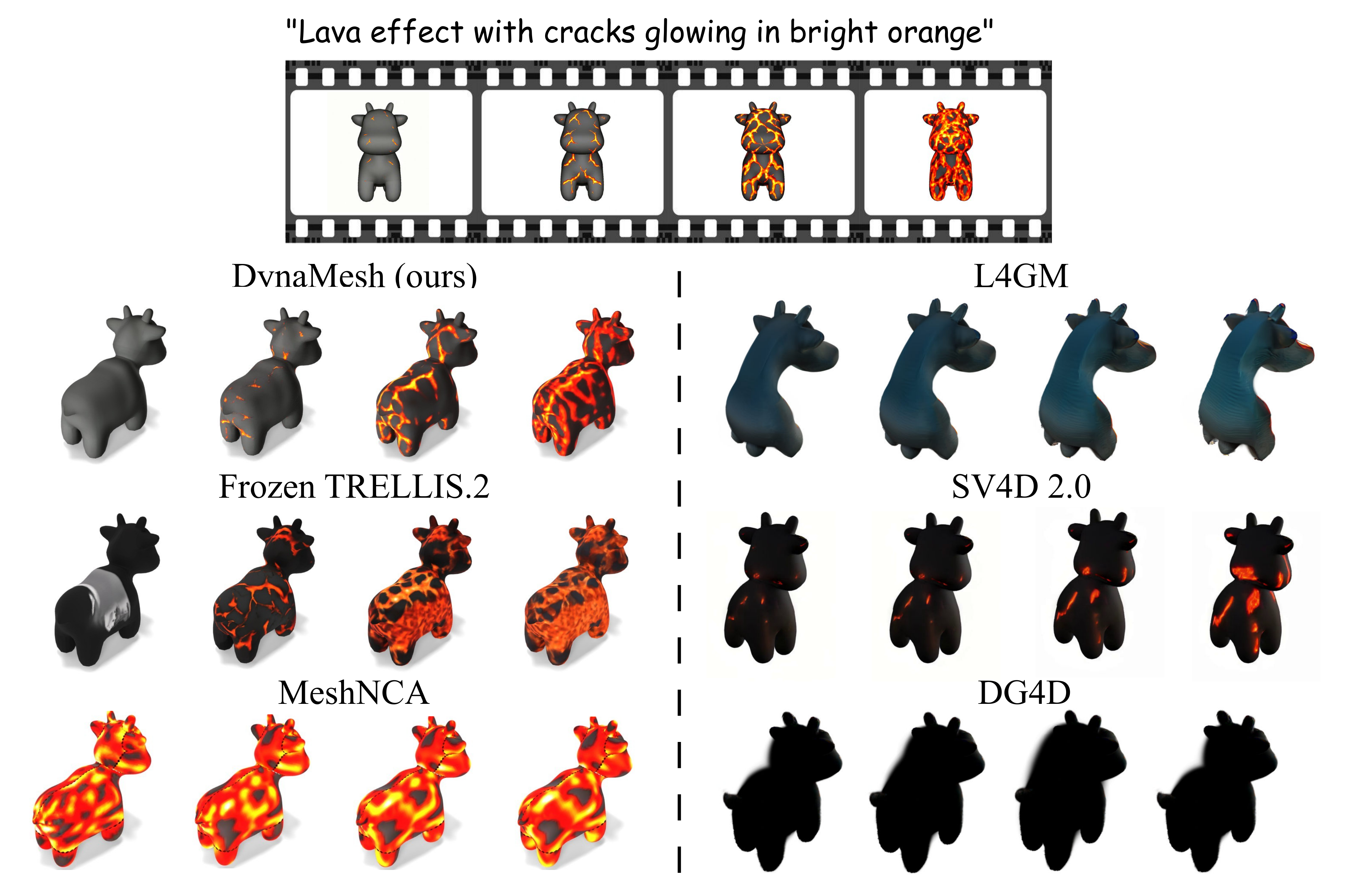}
\vspace{-2pt}
\caption{\textbf{Comparison at the second novel view.} All methods are shown from a camera behind the object, which none of them was supervised on, at the same four frames as \cref{fig:comparison_supp_a}. The baselines that construct their own geometry or operate in video space lose the effect. In contrast, our method presents the same texture it shows at the supervised view, since the effect is written onto the input mesh rather than onto a camera.}

\label{fig:comparison_supp_b}
\end{figure}

\smallskip
\noindent \textbf{Comparison.} \cref{fig:comparison_supp_a} compares every method on an object whose parts are thin and separated. The baselines that build their own geometry lose the leaves, so their effect is carried on a shape that is no longer the input one, whereas our method changes only appearance and keeps the underlying geometry the same.
\cref{fig:comparison_supp_b} repeats the comparison at the unseen camera behind the object, where the gap is widest.

\begin{figure}[!t]
    \centering
    \includegraphics[width=\columnwidth]{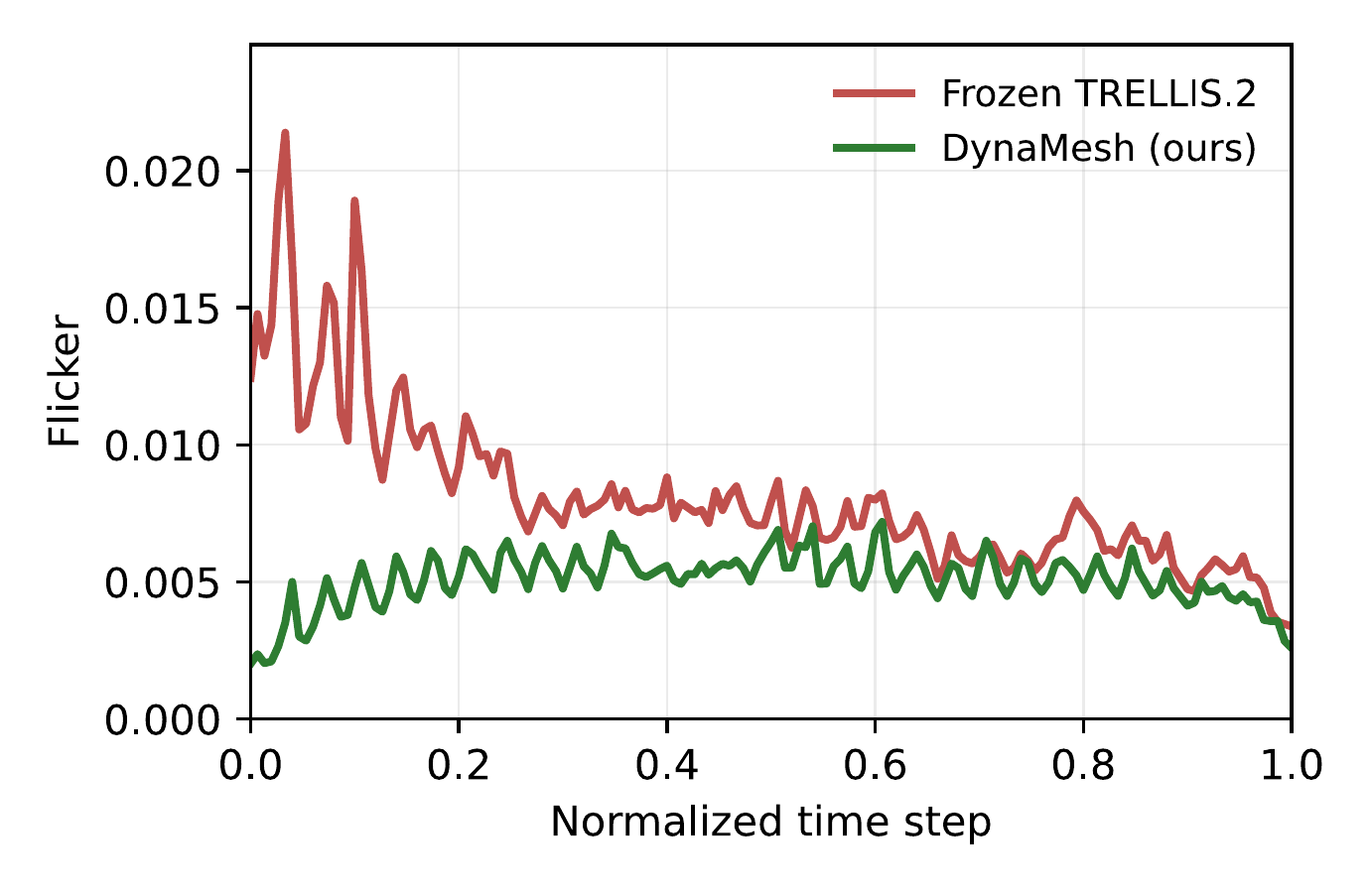}
    \vspace{-0.3cm}
 \caption{\textbf{Flicker over the clip.} We plot flicker at each time step for frozen TRELLIS.2 and for DynaMesh, over the 42 objects of \cref{tab:comparison}. The gap holds across the whole clip and is largest at the start, before the frozen texture settles.}

    \vspace{-0.3cm}
    \label{fig:flicker_supp}
\end{figure}

\smallskip
\noindent \textbf{Flicker over the clip.} Blending the conditioning of neighboring frames means consecutive frames receive nearly the same conditioning, so the texture has no instant at which it can jump.
 \cref{fig:flicker_supp} shows that this holds throughout the clip and not only on average. Our flicker stays below the frozen generator's at every time step.

\begin{figure*}[!t]
    \centering
    \includegraphics[width=\textwidth]{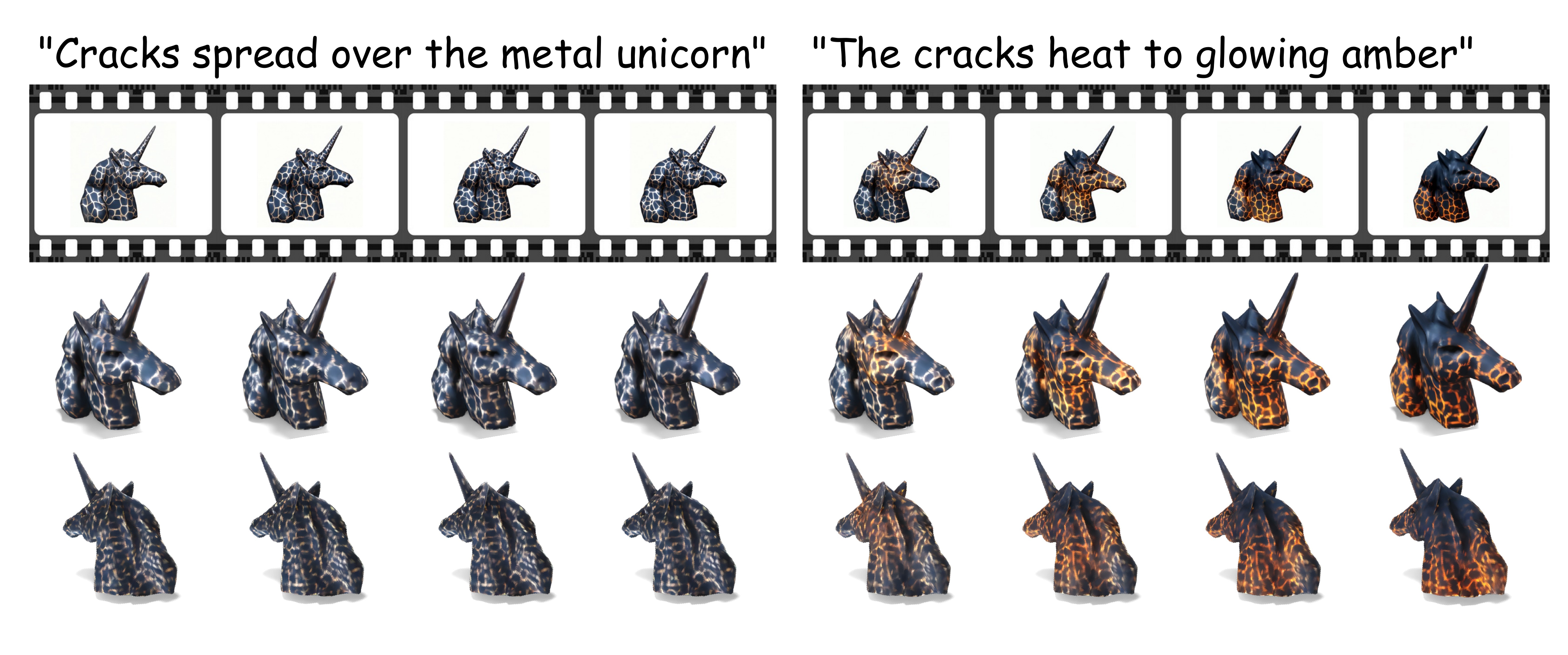}
    \vspace{-0.3cm}
    \caption{\textbf{Continuing an effect across clips.} A second reference video, generated from the last frame of the first, continues the same effect, so a progression that does not fit in one clip can be carried across several.}
    \vspace{-0.3cm}
    \label{fig:continued}
\end{figure*}

\smallskip
\noindent \textbf{Extended texture.} In \cref{fig:continued}, we fit the same object to a second reference video, generated from the last frame of the first. The crazing that forms on the bare metal in the first clip carries into the second, where the same cracks heat to amber.

\begin{table}[t!]
\centering
\small
\setlength{\tabcolsep}{4pt}
\begin{tabular}{ccc cc cc}
\toprule
\multicolumn{3}{c}{Attention} & \multicolumn{2}{c}{Unseen views} & \multicolumn{2}{c}{Fidelity \vs GT} \\
\cmidrule(lr){1-3}\cmidrule(lr){4-5}\cmidrule(lr){6-7}
Cross & Self & Temp. & Flicker $\downarrow$ & Accel. $\downarrow$ & PSNR $\uparrow$ & SSIM $\uparrow$ \\
\midrule
\textcolor{red!80!black}{\xmark} & \textcolor{red!80!black}{\xmark} & \textcolor{red!80!black}{\xmark} & 0.01818 & 0.02720 & 12.14 & 0.3035 \\
\textcolor{green!55!black}{\cmark} & \textcolor{red!80!black}{\xmark} & \textcolor{red!80!black}{\xmark} & 0.00727 & 0.01039 & 23.80 & 0.7542 \\
\textcolor{green!55!black}{\cmark} & \textcolor{green!55!black}{\cmark} & \textcolor{red!80!black}{\xmark} & 0.00702 & 0.00998 & 24.86 & 0.7970 \\
\textcolor{green!55!black}{\cmark} & \textcolor{green!55!black}{\cmark} & \textcolor{green!55!black}{\cmark} & \textbf{0.00499} & \textbf{0.00531} & \textbf{24.89} & \textbf{0.7980} \\
\bottomrule
\end{tabular}
\vspace{-2pt}
\caption{\textbf{Ablation study.} We turn on the components of our method one at a time and evaluate on the same 42 objects as \cref{tab:comparison}, with the same metrics. The first row is the frozen generator and the last row is our full method, which is best on every metric.}
\label{tab:ablation}
\end{table}

\begin{figure}[!t]
    \centering
    \includegraphics[width=\columnwidth]{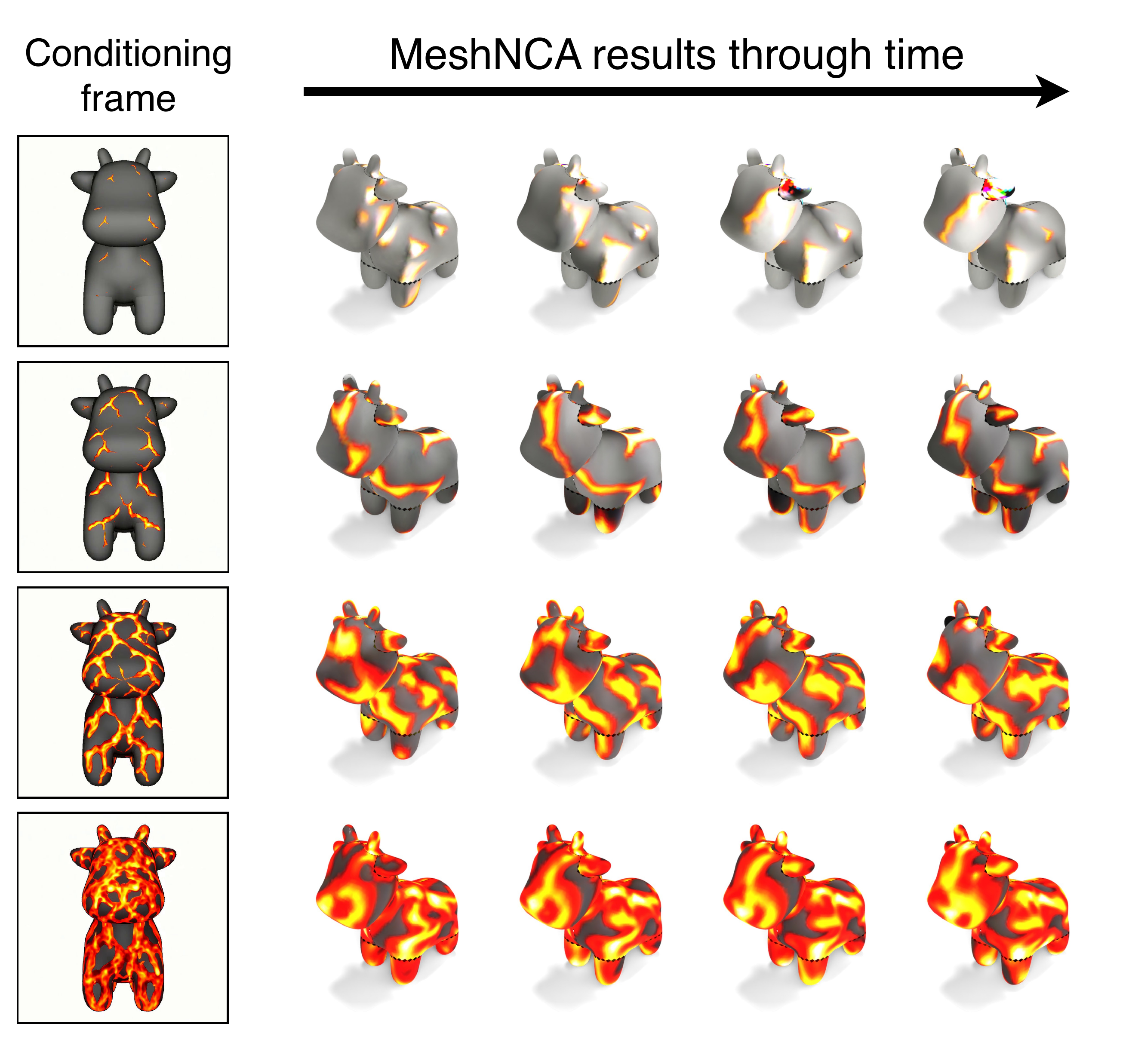}
    \vspace{-0.5cm}
   \caption{\textbf{MeshNCA conditioning.} Conditioned on a single frame, MeshNCA reproduces it and does not advance through time, so its result reflects the conditioning image rather than the progression of the effect.}

    \vspace{-0.3cm}
    \label{fig:meshnca_ablation}
\end{figure}

\section{LLM as a Judge} \label{sec:supp_judge}

\begin{table*}[t!]
\centering
\small
\setlength{\tabcolsep}{6pt}
\begin{tabular}{l cc cc cc}
\toprule
 & \multicolumn{2}{c}{Q1 effect fidelity (\%) $\uparrow$} & \multicolumn{2}{c}{Q2 temporal coherence $\uparrow$} & \multicolumn{2}{c}{Q3 geometry $\uparrow$} \\
\cmidrule(lr){2-3}\cmidrule(lr){4-5}\cmidrule(lr){6-7}
Method & Supervised & Unseen & Supervised & Unseen & Supervised & Unseen \\
\midrule
Frozen TRELLIS.2 & 85.9 & 85.6 & 3.66 & 3.71 & 4.90 & 4.86 \\
MeshNCA & 31.4 & 32.9 & 2.62 & 2.71 & 4.70 & 4.75 \\
SV4D 2.0 & 98.4 & 72.3 & 4.55 & 4.04 & \textbf{4.98} & 3.62 \\
DreamGaussian4D & 45.4 & 27.4 & 2.66 & 2.42 & 4.69 & 3.65 \\
L4GM & 98.9 & 60.3 & 4.64 & 3.75 & 4.90 & 3.47 \\
DynaMesh (ours) & \textbf{99.0} & \textbf{89.0} & \textbf{4.69} & \textbf{4.48} & 4.97 & \textbf{4.89} \\
\midrule
\textit{Reference video} & 98.9 & -- & 4.80 & -- & 4.92 & -- \\
\bottomrule
\end{tabular}
\vspace{-2pt}
\caption{\textbf{LLM as a judge.} We ask a vision-language model three questions about each rendered clip, at the supervised view and at the three unseen views, over all 42 objects with three repeats per clip. Effect fidelity is the percentage of five yes/no questions passed, while temporal coherence and geometry are rated from 1 (worst) to 5 (best). Our method scores best at the unseen views on all three questions, and the gap to the video-based baselines widens as the camera moves away from the supervised one.}

\label{tab:judge}
\end{table*}

\begin{table}[t!]
      \centering
      \small
      \setlength{\tabcolsep}{3pt}
      \begin{tabular}{llcccc}
        \toprule
        Effect & Shapes & Flick.$\downarrow$ & Accel.$\downarrow$ & Q1 (\%)$\uparrow$ & Q2$\uparrow$ \\     
        \midrule
        \multirow{2}{*}{Rorschach} & source & \textbf{0.004} & \textbf{0.005} & \textbf{100.0} & \textbf{5.00} \\ &
  unseen
    & 0.006 & 0.009 & \textbf{100.0} & 4.73 \\
        \midrule
        \multirow{2}{*}{Lava} & source & \textbf{0.007} & \textbf{0.005} & \textbf{86.7} & \textbf{5.00} \\ & unseen &
  \textbf{0.007} & 0.006 & 85.5 & 4.64 \\
        \bottomrule
      \end{tabular}
      \caption{\textbf{LLM as a Judge for Generalization.} Flicker, acceleration, and judge scores (Q1 effect fidelity, Q2 temporal coherence) for each adapter on the shape it was fit on and averaged over eleven unseen shapes. Effect fidelity is the percentage of five shape-neutral questions passed per effect, generated from the reference video, so the same questions score every shape. The transferred effects keep the temporal quality and the fidelity of the source result.}
  \label{tab:judge_gen}
  \end{table}

The reference video is captured from a single camera. Thus, to evaluate our results at the unseen views, where a target to score against does not exist, we use a vision-language model, Gemini 3.6 Flash~\cite{gemini2026}, as a judge. It answers three questions per rendered clip (\cref{tab:judge_protocol}): whether the effect matches the reference video, whether the clip is temporally coherent, and whether the geometry matches the input mesh. For effect fidelity, we generate five yes/no questions per object from its reference video, covering the presence of the effect, its appearance, where it sits on the object, how it progresses over the clip, and whether it stays within the object. The score is the fraction answered yes.

For geometry, we show the judge a gray render of the input mesh from the same camera and ask whether the shape is the same one, with color and texture excluded, so that a method producing a wrong but consistent geometry cannot pass. For temporal coherence, we ask the judge to rate flicker and popping between consecutive frames from 1 to 5. No reference is needed here, since both are visible within the clip itself. All 42 objects and six methods are judged at the four cameras, three times each. Each clip is rendered at $384 \times 384$ and uploaded at an explicit rate of 10 frames per second, since the API default of one frame per second hides flicker. Judgements are made at temperature 1.0. As a sanity check, the reference video answers its own fidelity questions at $98.9\%$.

\smallskip
\noindent \textbf{Results.} \Cref{tab:judge} reports the supervised view, the mean over the three unseen views, and the drop between them. Our method scores highest at the unseen views on all three questions. The video-based baselines fail to preserve the effect once the camera moves, losing up to $38.6$ points of effect fidelity and $1.43$ of geometry, whereas our method loses $10.0$ and $0.08$, demonstrating that the texture is bound to the surface and not to the supervised camera. Frozen TRELLIS.2 is view-consistent by construction and does not drop either, yet it reaches only $85.9\%$ effect fidelity at the supervised view against our $99.0\%$.

\smallskip
\noindent \textbf{Generalization.} We apply the same protocol to the transfer results of 11 unseen objects. The per-object fidelity questions would bias this comparison, since 44 of the 210 questions include specific parts about an object. We therefore regenerate the five questions once per effect with object parts excluded, so the same questions score the source shape and every unseen shape, and the judge is told that the two clips show different objects. \Cref{tab:judge_gen} reports the temporal metrics and the judge scores for each adapter on its source shape and averaged over the eleven unseen shapes. The transferred effects keep the temporal quality and the fidelity of the source result. We do not ask the geometry question here, since the mesh is encoded once and every transfer keeps the target geometry by construction.

\begin{table}[!b]
\centering
\small
\setlength{\tabcolsep}{4pt}
\resizebox{\columnwidth}{!}{%
\begin{tabular}{l l l}
\toprule
Question & Reference & Score \\
\midrule
Q1 effect fidelity      & reference video               & \% of 5 questions passed \\
Q2 temporal coherence   & none                          & 1 to 5 \\
Q3 geometry correctness & gray render of the input mesh & 1 to 5 \\
\bottomrule
\end{tabular}}
\vspace{-2pt}
\caption{\textbf{Judge protocol.} We ask three questions about every rendered clip, each scored against the reference its claim needs.}

\label{tab:judge_protocol}
\end{table}

\section{Component Ablation} \label{sec:supp_ablation}

\Cref{tab:ablation} switches the three components on one at a time over the 42 objects of \cref{tab:comparison}, measured exactly as in that table. The first row is the frozen generator and the last is our full method.

Most of the gain comes from the cross-attention adapter, which improves flicker by $60.0\%$ and increases the PSNR by $11.66$\,dB. Adding the self-attention adapter improves PSNR by another $+1.06$\,dB with flicker nearly unchanged. Temporal conditioning lowers flicker by $28.9\%$ and acceleration by $46.8\%$ with PSNR unchanged. This makes sense as PSNR is computed against per-frame targets and does not measure coherence, so its value is roughly unchanged on the last row.

\par
\begin{figure}[!t]
    \centering
    \includegraphics[width=\columnwidth]{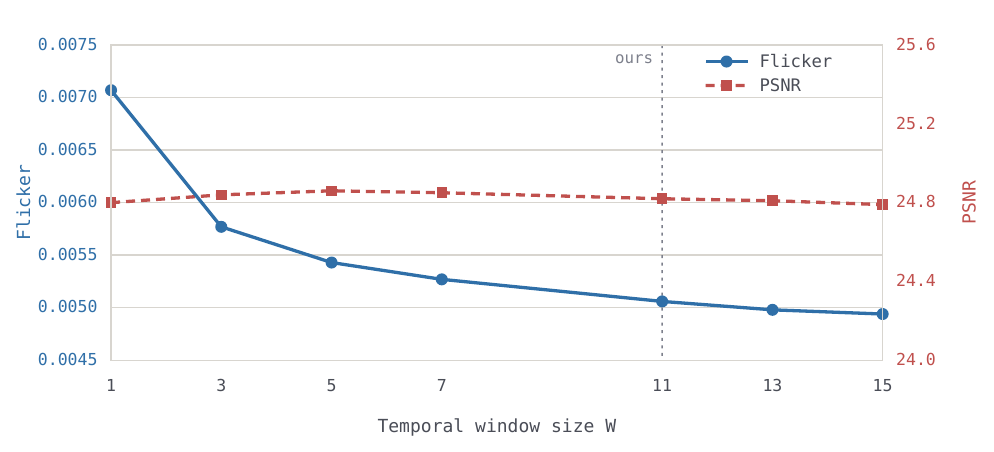}
    \vspace{-0.5cm}
    \caption{\textbf{Temporal window width.} Widening the conditioning window reduces flicker, but fidelity peaks at a width of five and declines beyond it, so we select eleven to balance the two.}
    \vspace{-0.3cm}
    \label{fig:window}
\end{figure}

\smallskip
\noindent \textbf{Window width.} The temporal component blends the conditioning of neighboring frames, and \cref{fig:window} sweeps how many. Widening the window keeps reducing flicker. Fidelity peaks at a width of five and only slightly falls after, so we choose eleven, where the flicker gain is large, and the fidelity cost is negligible.

\par\smallskip

\noindent \textbf{Temporal operator ablation.} To examine whether a learned operator improves on the parameter-free blend of \cref{eq:blend}, we replace it with two adapted attention operators at the same eleven-frame window, everything else held fixed: \emph{temporal attention}, which duplicates the cross-attention of \cref{eq:ca} into a second adapted pass where token $i$ attends only to token $i$ across the window, and \emph{spatio-temporal attention}, which gives that pass the whole window at once, so a single softmax runs over all $11N$ tokens. Neither improves on the blend (\cref{tab:tempattn}): both flicker and accelerate more on all 42 objects, and match it on PSNR and SSIM to within $0.07$\,dB and $0.002$.

\begin{table}[t]
\vspace{10pt}
\centering
\small
\setlength{\tabcolsep}{4pt}
\begin{tabular}{lcccc}
\toprule
& \multicolumn{2}{c}{Temporal} & \multicolumn{2}{c}{Reconstruction} \\
\cmidrule(lr){2-3}\cmidrule(lr){4-5}
Temporal operator & Flicker $\downarrow$ & Accel. $\downarrow$ & PSNR $\uparrow$ & SSIM $\uparrow$ \\
\midrule
Temporal attention     & 0.00680 & 0.00921 & \textbf{24.89} & \textbf{0.7171} \\
Spatio-temporal attn.  & 0.00683 & 0.00924 & 24.86 & 0.7165 \\
Parameter-free (ours)  & \textbf{0.00506} & \textbf{0.00474} & 24.82 & 0.7149 \\
\bottomrule
\end{tabular}
\vspace{-2pt}
\caption{\textbf{Temporal operator ablation.} We compare two learned temporal operators against the parameter-free blend, at the same eleven-frame window with everything else held fixed. Metrics in the table are computed on the texture rather than on the renders, so they differ from \cref{tab:ablation}. The blend has lower flicker and acceleration on all 42 objects, at a cost of $0.07$\,dB in PSNR and $0.002$ in SSIM.}
\label{tab:tempattn}
\end{table}

\section{Settings} \label{sec:supp_impl} 

\Cref{tab:supp_settings} lists every value we fixed and the training details.

\begin{table}[t!]
\centering
\small
\setlength{\tabcolsep}{5pt}
\begin{tabular}{lll}
\toprule
Stage & Setting & Value \\
\midrule
\multirow{3}{*}{Reference video}
 & model            & Kling 3.0~\cite{kling2026} \\
 & clip length      & 121 to 150 frames \\
 & training res.    & $960^2$ \\
\midrule
\multirow{4}{*}{Generator}
 & backbone         & TRELLIS.2~\cite{xiang2025trellis2} \\
 & image encoder    & DINOv3 ViT-L/16~\cite{simeoni2025dinov3} \\
 & flow blocks      & 30 \\
 & ODE steps        & 12 \\
\midrule
\multirow{2}{*}{Blending}
 & window           & $[t{-}5,\,\ldots,\,t,\,\ldots,\,t{+}5]$ \\
 & number of frames           & 11 \\

\midrule
\multirow{7}{*}{Adaptation}
 & rank $r$         & 4 \\
 & scaling $\alpha$ & 4 \\
 & adapted matrices & 150 \\
 & trainable params & $2{,}334{,}720$ \\
 & fraction frozen  & $99.94\%$ \\
 & epochs           & 30 \\
 & seed             & 42 \\
\midrule
\multirow{2}{*}{Loss}
 & reconstruction   & $L_1$ on color \\
 & perceptual       & LPIPS, weight $0.1$ \\
\midrule
\multirow{2}{*}{Hardware}
 & GPU              & one Nvidia A40 \\
 & peak memory      & 17.6\,GiB \\
\bottomrule
\end{tabular}
\vspace{-2pt}
\caption{\textbf{Implementation details.} We list the generator, the blending window, the adapter, the loss, and the hardware used to produce every result in the paper. All objects are fit with these same values.}

\label{tab:supp_settings}
\end{table}

\smallskip
\noindent \textbf{Prompt format.} Every prompt has two parts. The first is a fixed preamble that states the object must stay in place and keep its size, orientation, and shape for the whole video. The second names the effect in a short phrase, for example \textit{rot spreads and darkens}. Only the second part changes between objects. Here is an example of a prompt for the lava effect on the spot mesh: 
\begin{quote}\small\ttfamily
"Static tripod framing; the camera is locked. No pan, tilt, roll, dolly, push-in, pull-out, zoom, reframing, or shake. The cow is a single rigid statue: no breathing, swaying, head turn, tail flick, ear movement, limb motion, or micro-jitter; its pose, scale, position, and silhouette stay pixel-identical in every frame. The background is flat pure white and never changes; no embers, no heat haze, no glow spill, no particles anywhere outside the object. The effect lives strictly on the surface and never crosses the silhouette edge; nothing drips off, nothing is emitted into the air, nothing lights the white background. The only thing that changes is the colour of the surface itself. Molten lava glows inside deep cracks running across the cow's body, orange-red magma creeping and pulsing along the fissures, flaring to yellow-white at random points then cooling to black crust, new fissures opening unpredictably across the flank, head and legs."
\end{quote}

\smallskip
\noindent \textbf{Training and Inference Runtime.} Fitting the adapter for one effect takes 2.8 hours on average on a single NVIDIA A40, measured over 42 objects at 30 epochs and 150 frames. Inference takes 2.5 seconds per frame on average, and a 150-frame sequence completes in eight minutes. On the same hardware, MeshNCA fits in 16 minutes per scene on average and renders in one minute. L4GM takes 3 minutes per scene on average, SV4D 2.0 takes 11 minutes, and DreamGaussian4D takes 30 minutes.

\smallskip
\noindent \textbf{Optimization.} We use the generator's own flow sampler, which denoises in 12 steps. Each training step picks one video frame and one of the 12 steps at random. We run the sampler up to that step without gradients, then take one more step with gradients enabled and get its prediction of the clean sample. We decode and render it to compute the loss against the target. Only the adapter weights are updated, and we train them with AdamW at a learning rate of $10^{-4}$ for 30 epochs over 150 frames.

\smallskip
\noindent \textbf{Adapted projections.} \label{sec:supp_proj}
Let $h_t \in \mathbb{R}^{N \times c}$ be the appearance features of the $N$ occupied voxels of $\Scal$ at frame $t$, with channel width $c$, and $\tilde{z}_t \in \mathbb{R}^{M \times c}$ for the $M$ blended conditioning tokens of \cref{eq:blend}. The two adapted attention formulas are
\begin{equation} \label{eq:ca}
    \mathrm{CA}(h_t, \tilde{z}_t) = \mathrm{softmax}\!\left( \frac{(h_t W'_{q})(\tilde{z}_t W'_{k})^{\top}}{\sqrt{d}} \right) (\tilde{z}_t W'_{v}) \, W'_{o} ,
\end{equation}
\begin{equation} \label{eq:sa}
    \mathrm{SA}(h_t) = \mathrm{softmax}\!\left( \frac{(h_t W'_{q})(h_t W'_{k})^{\top}}{\sqrt{d}} \right) (h_t W'_{v}) \, W'_{o} ,
\end{equation}
where $d$ is the attention dimension and each $W'$ has its own adapter from \cref{eq:lora}. Cross-attention writes the frame's conditioning onto the voxels and self-attention passes the effect details between the voxels.

\section{Cameras and Metrics} \label{sec:supp_metrics}

Every method is rendered at $512^2$ from four cameras. The supervised view is the camera the reference video was generated from. The three others are unseen during fitting and sit at azimuth and elevation $(45^{\circ}, 25^{\circ})$, $(135^{\circ}, -20^{\circ})$, and $(225^{\circ}, 30^{\circ})$.

Flicker and acceleration are computed over each method's own foreground, as defined in the main paper. This is because two of the baselines construct their own geometry for the effect (L4GM~\cite{ren2024l4gm} and DG4D~\cite{ren2023dreamgaussian4d}).

\smallskip
\noindent \textbf{PSNR.} Let $R_t$ be the render at time $t$, $I_t$ the reference frame, and $\Omega_t$ the silhouette of $I_t$. Pixel values lie in $[0, L]$ with $L = 1$. The squared error inside the silhouette is
\begin{equation} \label{eq:mse}
    \mathrm{MSE}_t \;=\; \frac{1}{3|\Omega_t|} \sum_{p \in \Omega_t} \big( R_t(p) - I_t(p) \big)^2 ,
\end{equation}
and we report the mean over frames,
\begin{equation} \label{eq:psnr}
    \mathrm{PSNR} \;=\; \frac{1}{T} \sum_{t=1}^{T} 10 \log_{10} \frac{L^2}{\mathrm{MSE}_t} .
\end{equation}

\smallskip
\noindent \textbf{SSIM.} For a pair of co-located windows $x$ in $R_t$ and $y$ in $I_t$, with means $\mu_x, \mu_y$, variances $\sigma_x^2, \sigma_y^2$ and covariance $\sigma_{xy}$,
\begin{equation} \label{eq:ssim}
    \mathrm{SSIM}(x,y) \;=\; \frac{(2\mu_x\mu_y + C_1)(2\sigma_{xy} + C_2)}{(\mu_x^2 + \mu_y^2 + C_1)(\sigma_x^2 + \sigma_y^2 + C_2)} ,
\end{equation}
where $C_1 = (k_1 L)^2$, $C_2 = (k_2 L)^2$, $k_1 = 0.01$ and $k_2 = 0.03$. Windows are Gaussian, $11 \times 11$ with $\sigma = 1.5$. We average over the windows centered inside $\Omega_t$ and then over frames.

Both quantities are measured only at the supervised view. The reference clip is a video from a single camera, so the three unseen views have nothing to score against.

\section{Baseline Settings} \label{sec:supp_baselines}

\noindent \textbf{SV4D 2.0.} Four-view model, 50 sampling steps, seed 23, encoding chunk 8, decoding chunk 4, background removal off. Inputs are uniformly subsampled to 21 frames at $576^2$, which is the model's cap.

We moved its cameras onto our rig so the comparison is angle matched, giving azimuth $0^{\circ}$, $45^{\circ}$, $135^{\circ}$, $180^{\circ}$, $225^{\circ}$ at elevation $0^{\circ}$, $25^{\circ}$, $-20^{\circ}$, $0^{\circ}$, $30^{\circ}$. SV4D 2.0 is trained on azimuth $0^{\circ}$, $60^{\circ}$, $120^{\circ}$, $180^{\circ}$, $240^{\circ}$ at elevation $0^{\circ}$, so our angles are inside its API and outside its training distribution, which may lower its quality. On its native grid, only azimuth $0^{\circ}$ overlaps ours, and SV4D 2.0 conditions on that view. Its supervised view is that conditioning view re-encoded, and we mark its fidelity numbers in \cref{tab:comparison} accordingly.

\smallskip
\noindent \textbf{L4GM and DG4D.} Both receive the reference video and build their own geometry. DG4D outputs 32 frames, matched onto the 21 shared instants.

\smallskip
\noindent \textbf{MeshNCA.} We give MeshNCA the same mesh as the input for our pipeline. Its condition is an image with no video or text input, so we give it the last frame of the reference video. To check that this choice does not determine its result, we refit it on each of the four frames of the reference video and roll every model out through time. As shown in \cref{fig:meshnca_ablation}, the texture does not advance across a row, so the desired temporal effect cannot be achieved regardless of the input.

\end{document}